\pdfoutput=1

\documentclass[11pt]{article}

\usepackage[preprint]{acl}

\usepackage{times}
\usepackage{latexsym}

\usepackage[T1]{fontenc}

\usepackage[utf8]{inputenc}

\usepackage{microtype}

\usepackage{inconsolata}

\usepackage{graphicx}

\usepackage{array}
\usepackage{multirow}
\usepackage{CJKutf8}
\usepackage{longtable}
\usepackage{float}
\usepackage{tabularx}
\usepackage[most]{tcolorbox}
\usepackage{booktabs}
\usepackage{makecell}

\usepackage{graphicx}
\usepackage{subcaption}

\usepackage{amsmath}  

\title{C-ReD: A Comprehensive Chinese Benchmark for AI-Generated \\Text Detection Derived from Real-World Prompts}

\author{Chenxi Qing$^{1}$\thanks{Equal contribution.}~, \quad
Junxi Wu$^{2,1}$\footnote[1]{}~, \quad
Zheng Liu$^{1}$\footnote[1]{}~, \quad
Yixiang Qiu$^{1}$, \\
\textbf{Hongyao Yu$^{1}$, \quad
Bin Chen$^{3,4}$\thanks{Corresponding author.}~, \quad
Hao Wu$^{1,5}$\footnote[2]{}~, \quad Shu-Tao Xia$^{1,4}$}\\
$^{1}$ Tsinghua University \quad $^{2}$ Nankai University \\ 
$^{3}$ Harbin Institute of Technology, Shenzhen \\
$^{4}$ Peng Cheng Laboratory \quad
$^{5}$ Shannon InfoTech\\
\texttt{qcx25@mails.tsinghua.edu.cn}, \quad
\texttt{chenbin2021@hit.edu.cn}\\
}

\begin{document}
\maketitle

\begin{abstract}
Recently, large language models (LLMs) are capable of generating highly fluent textual content. While they offer significant convenience to humans, they also introduce various risks, like phishing and academic dishonesty. Numerous research efforts have been dedicated to developing algorithms for detecting AI-generated text and constructing relevant datasets. However, in the domain of Chinese corpora, challenges remain, including limited model diversity and data homogeneity. To address these issues, we propose \textbf{C-ReD}: a comprehensive \textbf{C}hinese \textbf{Re}al-prompt AI-generated text \textbf{D}etection benchmark. Experiments demonstrate that C-ReD not only enables reliable in-domain detection but also supports strong generalization to unseen LLMs and external Chinese datasets—addressing critical gaps in model diversity, domain coverage, and prompt realism that have limited prior Chinese detection benchmarks. We release our resources at \url{https://github.com/HeraldofLight/C-ReD}.

\end{abstract}

\section{Introduction}

Large language models (LLMs) have quickly become integral to everyday life and professional workflows. Leading models such as ChatGPT~\cite{brown2020language} and DeepSeek~\cite{guo2025deepseek} generate highly fluent and contextually relevant responses, greatly boosting productivity and user experience. However, this same capability can be exploited for malicious or unethical purposes~\cite{fang2025your}—such as phishing, academic dishonesty~\cite{tang2024science}, plagiarism~\cite{lee2023language}, and the dissemination of disinformation~\cite{mitchell2023detectgpt}. Compounding the problem, human readers often find it difficult to reliably tell AI-generated text apart from human-written content~\cite{mitchell2023detectgpt}, which further exacerbates these risks.

To address this challenge, substantial research efforts have focused on developing detectors for AI-generated text~\cite{solaiman2019release,mitchell2023detectgpt,yang2024dna,bao2024fast,chen2025imitate}. Concurrently, several benchmark datasets have been released to support detector training and evaluation~\cite{uchendu2021turingbench,li2023mage,he2024mgtbench,wu2024detectrl}. However, these datasets predominantly consist of English corpora. 

Chinese presents unique challenges for detection due to its complex word segmentation~\cite{tsang2025corpus}, context-sensitive semantics, and abundant use of cultural idioms, metaphors, and informal abbreviations. These linguistic properties render direct adaptation of English-centric methods ineffective and underscore the need for native Chinese detection benchmarks. Unfortunately, prior Chinese datasets suffer from three key limitations:  
(1) \textbf{Limited model diversity}: most rely solely on ChatGPT~\cite{macko2023multitude,wang2023m4}, overlooking widely adopted domestic Chinese LLMs;  
(2) \textbf{Data homogeneity}: texts are often restricted to simple QA formats~\cite{guo2023close}, lacking representation from professional domains such as journalism or academic writing;
(3) \textbf{Unrealistic prompt design}: failing to reflect real-word application scenarios of LLMs.

To bridge these gaps, we introduce \textbf{C-ReD}: a comprehensive \textbf{C}hinese \textbf{Re}al-prompt AI-generated text \textbf{D}etection benchmark. C-ReD comprises Chinese human-written texts curated from five distinct domains, paired with AI-generated counterparts produced by nine LLMs—including five leading Chinese domestic models—under five carefully designed, real-world-inspired prompt types.
\begin{figure*}
    \centering
    \includegraphics[width=\linewidth]{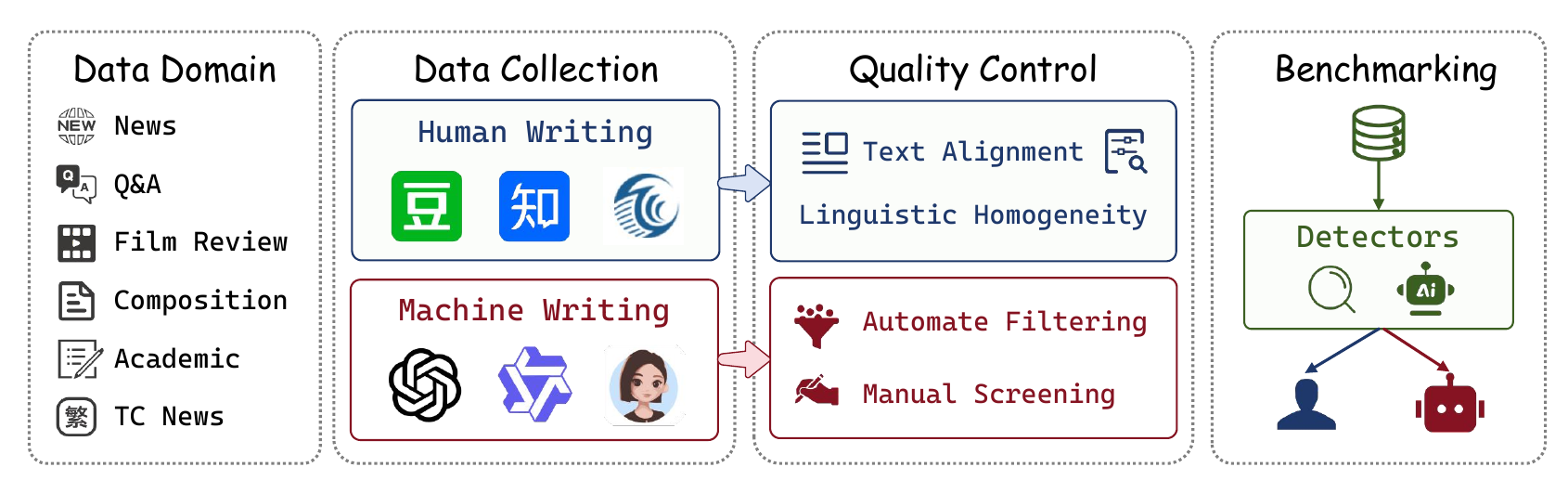}
    \caption{The overview of C-ReD.}
    \label{fig:pipeline}
\end{figure*}

Our analysis using C-ReD reveals that detection performance varies substantially across domains and generators, with fluent, reasoning-intensive models like Deepseek-R1 being particularly challenging to detect. Fine-tuning on C-ReD not only boosts in-domain accuracy but also enables strong generalization to unseen models and external datasets, demonstrating its effectiveness as a representative and scalable foundation for Chinese AI-generated text detection.

\section{Related Work}

\subsection{Detection Methods}
\textbf{Supervised Methods.} The goal of supervised methods is training a classifier, which typically leverages neural representations from pre-trained models. RoBERTa~\cite{liu2019roberta}, in particular, has been widely adopted for this purpose. Studies fine-tune it on labeled datasets, use its contextual embeddings to train binary classifiers \cite{solaiman2019release,guo2023close}, and leverage its semantic capture ability to achieve effective detection in specific domains. 
ImBD \cite{chen2025imitate} identified machine-revised text by imitating the machine-style token distribution by style preference optimization.
Recent works incorporate token-level probability distributions with pre-trained model embeddings \cite{verma2024ghostbuster,shi2024ten}, utilizing both semantic and statistical information effectively. 
MoSEs \cite{wu2025moses} introduces a flexible framework that models profession-specific writing styles and addresses the limitation of static threshold by employing conditional threshold estimation to enable adaptive detection across diverse domains.
The performance of supervised methods is fundamentally based on the quality and representativeness of training data. High-quality datasets, with diverse text types, comprehensive source models, and accurate labels, can alleviate overfitting and improve generalization.

\textbf{Zero-shot Methods.} Existing zero-shot detectors mostly rely on constructing statistical features through pre-trained large language models, without requiring task-specific training. Early approaches rely on token-level likelihood–based metrics such as entropy~\cite{gehrmann2019gltr}, perplexity~\cite{lavergne2008detecting}, log-likelihood~\cite{solaiman2019release}, and top-$k$ probability buckets~\cite{gehrmann2019gltr}.
DetectGPT \cite{mitchell2023detectgpt} pioneered the paradigm of comparing perturbed texts with original ones for detection, inspiring related methods such as DetectNPR \cite{su2023detectllm} and DNA-GPT \cite{yang2024dna}. However, these methods suffer from high time costs. Fast-DetectGPT \cite{bao2024fast} improved efficiency via conditional probability curvature, expanding the application scope of training-free detection methods. Lastde \cite{xu2025trainingfree} introduced time series analysis for better accuracy and DNA-DetectLLM\cite{zhu2025dna} introduced the mutation-repair paradigm. Addressing the limitations of prior likelihood-based approaches, LAPD\cite{wu2026alignment} makes a significant theoretical and practical contribution by formalizing the alignment imprint as a measurable statistic. This novel formulation not only provides a standardized and efficient approach for detection but also offers a rigorous framework for understanding the statistical signatures left by model alignment processes. Note that zero-shot methods typically utilize labeled data to perform probability calibration.

\subsection{Related Datasets}
\textbf{AI-generated Text Benchmarks.} A growing body of research has addressed the challenges posed by large language model (LLM)-generated text, leading to the development of several benchmark datasets. TuringBench~\cite{uchendu2021turingbench} pioneered this direction by introducing a dataset consisting of 200k human-written and AI-generated news articles produced by 19 distinct LLMs. However, it is limited by its reliance on models no more advanced than GPT-3 and by the narrow diversity of its data sources. Building on this foundation, MAGE~\cite{li2023mage} presents a large-scale testbed comprising outputs from 27 LLMs across seven diverse writing tasks. MGTBench~\cite{he2024mgtbench} further advances the field by offering a unified framework for both detection and attribution of machine-generated text, enabling systematic evaluation of various methods across multiple datasets. To better reflect real-world deployment scenarios, DetectRL~\cite{wu2024detectrl} introduces four evaluation tasks covering high-risk domains such as social media and news writing—contexts particularly susceptible to synthetic text abuse. Despite these advances, existing benchmarks remain predominantly English-centric and often lack coverage of Chinese-language LLMs, highlighting a critical gap for non-English detection research.

\textbf{Chinese Corpora.}
Several studies have included Chinese AI-generated text in their benchmarks, yet significant limitations remain. HC3 \cite{guo2023close} provides 40k question–answer pairs in both English and Chinese, covering diverse domains such as computer science, law, and medicine. However, it is restricted to a single QA format and relies solely on ChatGPT for generation. MULTITuDE~\cite{macko2023multitude} is a multilingual detection benchmark encompassing 74k AI-generated texts across 11 languages—including Chinese—produced by eight LLMs. Notably, the Chinese subset was used only for evaluation and excluded from model training, limiting its utility for developing Chinese-specific detectors. M4~\cite{wang2023m4} introduces a multi-generator, multi-domain, and multilingual benchmark that includes Chinese data sourced from Baike and Web QA. Nevertheless, its Chinese AI-generated texts were produced exclusively by ChatGPT and GPT-3.5, omitting widely deployed domestic Chinese LLMs. Collectively, these Chinese datasets suffer from narrow domain coverage, homogeneous text formats, and limited model diversity—particularly the absence of state-of-the-art Chinese foundation models. Coupled with their relatively small scale and constrained scenario representation, these shortcomings render existing resources insufficient for robust, real-world Chinese AI-generated text detection.

\section{C-ReD Dataset}

\subsection{Data Sourcing}
We collected human-written texts from five domains where LLMs are commonly deployed. The data include:  
(1) \textbf{News}: 3,000 articles from THUCNews~\cite{li2006comparison}, evenly distributed across five categories (sports, politics, finance, entertainment, education);  
(2) \textbf{Q\&A}: 2,956 user answers from Zhihu via Zhihu-KOL~\cite{zhihukol};  
(3) \textbf{Film Review}: 2,960 user reviews from Douban, sourced from ChineseNlpCorpus~\cite{AesopChowChineseNLPCorpus};  
(4) \textbf{Composition}: 1,081 model essays from China’s National College Entrance Examination, collected via web scraping;  
(5) \textbf{Academic Writing}: 500 Chinese papers (title, keword, abstract, introduction) from ChinaXiv~\cite{chinaxiv}. 
For news domain, we also consider Traditional Chinese (TC) news for extra dataset and collect 2,000 articles from News-Collection-Zhtw~\cite{news-collection-zhtw}, covering four categories (article, tech, science, daily-weekly). Further information on data sources and statistics is provided in Appendix~\ref{appendix:dataset_construction data_source}.

\subsection{Model Sets}
We adopt a diverse set of representative LLMs to generate AI-written texts. To better reflect real-world Chinese AI writing scenarios, we include advanced models from both Chinese and international providers. Our model set includes nine LLMs: \textbf{OpenAI} (\texttt{gpt-3.5-turbo}, \texttt{gpt-4o})~\cite{gpt-3.5-turbo,hurst2024gpt}, \textbf{Google} (\texttt{Gemini-2.5-Flash})~\cite{comanici2025gemini}, \textbf{Anthropic} (\texttt{Claude-3.5-Haiku})~\cite{anthropic2024model}, \textbf{Deepseek} (\texttt{Deepseek-V3}, \texttt{Deepseek-R1})~\cite{liu2024deepseek,guo2025deepseek}—with Deepseek-R1 employed in chain-of-thought mode—, \textbf{Qwen} (\texttt{Qwen2.5}, \texttt{Qwen3})~\cite{qwen2.5,qwen3}, and \textbf{Doubao} (\texttt{Doubao-1.5-Pro})~\cite{doubao1.5pro}. All models are accessed exclusively via APIs, this black-box setting closely mirrors real-world usage but also increases the difficulty of detection, further details are provided in the Appendix~\ref{appendix:dataset_construction generative_llms}.

\subsection{Prompt Design}
To enhance the alignment between LLM-generated texts and real-world applications, we designed five domain-specific prompt strategies. \textbf{News}: we curated category-specific prompt templates with LLM assistance and combined them with real news headlines to simulate AI-assisted news writing. \textbf{Q\&A}: we first created general-purpose prompt templates and then specialized them by injecting domain-specific questions, mimicking real-world QA systems. \textbf{Film Review}: we developed structured review guidelines using LLMs and paired them with film titles to form final prompts, reflecting practical AI-powered critique generation. \textbf{Composition}: we directly utilized descriptions from Gaokao along with specific topics to construct the final prompts. \textbf{Academic Writing}: for introduction generation, we fused paper titles, keywords, and abstracts with instructional prompts. Separately, we also designed prompts to generate abstracts solely from titles and keywords. 
\textbf{TC News}: analogous to the News setup, we adapted similar template design methodology to Traditional Chinese journalism.
Sample prompts for each domain are provided in Appendix~\ref{appendix:dataset_construction prompt_design_and_examples}.

\begin{table}[h]
\small
\setlength{\tabcolsep}{4pt}
\caption{Core schema shared by all samples in the dataset. Fields below the second midrule are only present for AI-generated samples.}
\label{tab:csv_schema}
\begin{tabularx}{\linewidth}{@{}l l X@{}}
\toprule
\textbf{Field} & \textbf{Type} & \textbf{Description} \\
\midrule
\texttt{id}           & int & Unique sample identifier \\
\texttt{text}         & str & Human-written or AI-generated text \\
\texttt{label}        & int & 1 = human, 0 = AI \\
\texttt{type}         & int & Encoded text domain\\
\texttt{length}       & int & Character count \\
\texttt{attribution}  & str & Source of the text: model name for AI outputs, or \texttt{human} for human-written samples \\
\midrule
\texttt{original\_id} & int & ID of the original human-written sample used for AI generation (AI only) \\
\texttt{prompt}       & str & The exact prompt provided to the language model to generate this AI text (AI only) \\
\bottomrule
\end{tabularx}
\end{table}

\subsection{Quality Control}
To ensure high-quality and consistent data, we performed systematic preprocessing on both human-written and AI-generated texts.

\textbf{Human-written Text.} Human-authored content—sourced from diverse platforms—often contained extraneous elements such as metadata (e.g., headlines, author names, source URLs), non-textual components (e.g., figures, tables, citations), and formatting artifacts introduced during web scraping or PDF conversion. To align these with the clean, paragraph-level format of machine-generated outputs, we removed all such noise, standardized text structure into single coherent paragraphs, and applied length constraints appropriate to each genre. Additionally, we filtered out samples with excessive English content or inconsistent language use to maintain linguistic homogeneity. Full details of the cleaning protocols for each data type are provided in Appendix~\ref{appendix:dataset_construction data_source}.

\textbf{AI-generated Text.} To ensure the quality and consistency of AI-generated text, we implemented length constraints in the prompt templates to control output within predefined ranges for each domain. We adopted a dual-measure quality assurance approach: (1) real-time automated filtering that monitors Chinese character ratio, repetitive character sequences, text length, empty outputs, and factual inconsistencies, while also stripping extraneous formatting such as Markdown syntax, section headings, bullet points, and other non-paragraph elements; (2) rigorous manual screening by domain experts to identify and remove low-quality or anomalous samples. Specific parameter values for these quality controls are provided in Appendix~\ref{appendix:dataset_construction ai_generation_constrains}.

All texts (human-written and AI-generated) were normalized into plain, single-paragraph format (more details for text length in Appendix~\ref{appendix:text_length_statistics}). The final dataset is stored as a CSV file with a unified core schema (Table~\ref{tab:csv_schema}); domain-specific fields are detailed in Appendix~\ref{appendix:dataset_construction_csv_shcema}.

\subsection{Statistics}
\begin{figure}[htbp]
  \includegraphics[width=1\linewidth]{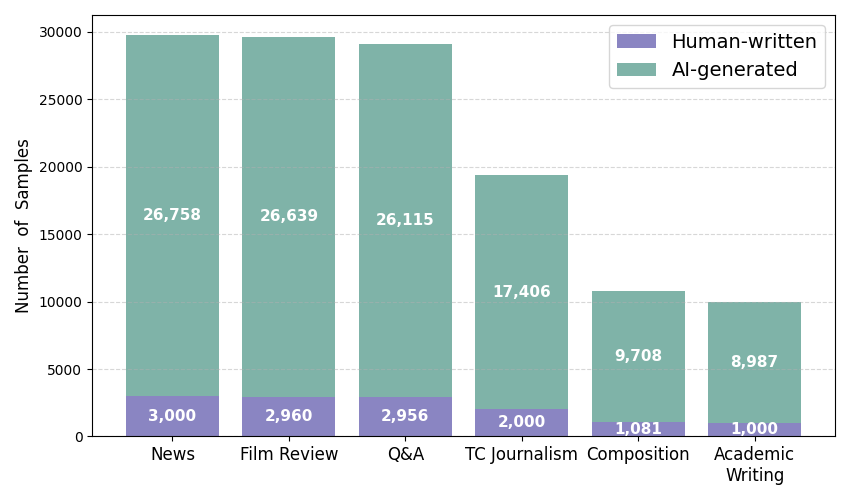}
  \caption{Distribution of samples in C-ReD.}
  \label{fig:data_distribution}
\end{figure}

As shown in Figure~\ref{fig:data_distribution},our dataset comprises a total of 128,610 texts, including 12,997 human-written and  115,613 AI-generated samples across five domains (including extra dataset). The full distribution across domains and large language models is provided in Appendix~\ref{appendix:dataset_construction statistics}. Examples of C-ReD are included in Appendix~\ref{appendix:examples of C-ReD}.

\section{Experimental Setup}
\subsection{Detectors}
Using C-ReD, we conduct a comprehensive evaluation of a diverse set of state-of-the-art AI-generated text detection methods, spanning both traditional paradigms—zero-shot and supervised methods. In particular, we further investigate the emerging paradigm of leveraging Large Language Models (LLMs) themselves as detectors, exploring their potential to identify AI-generated text without task-specific training.

\subsubsection{Zero-shot Detectors}
We evaluate a range of zero-shot detection methods. These include likelihood-based metrics: \textbf{Log-Likelihood}~\cite{solaiman2019release},  \textbf{Entropy}~\cite{gehrmann2019gltr},  and \textbf{Log-Rank}~\cite{mitchell2023detectgpt}. We also assess \textbf{LRR}~\cite{su2023detectllm}, which combines log-likelihood and log-rank into a normalized score; \textbf{Fast-DetectGPT}~\cite{bao2024fast}, which leverages conditional probability curvature; \textbf{Lastde} and \textbf{Lastde++}~\cite{xu2025trainingfree}, which analyze likelihood sequences as time-series signals; \textbf{DNA-DetectLLM}~\cite{zhu2025dna}, which enhances robustness through a mutation–repair mechanism; and \textbf{LAPD}~\cite{wu2026alignment}, which utilizes standardized information-weighted statistic based on alignment imprint. Full descriptions and details are provided in Appendix~\ref{appendix:detectors_zero-shot_detectors}.

\subsubsection{Supervised Detectors}
 We evaluate several representative methods, including the \textbf{OpenAI Detector}~\cite{solaiman2019release}, a RoBERTa-based classifier trained on GPT outputs; \textbf{RADAR}~\cite{hu2023radar},  which jointly trains an AI-generated text detector by adversarial learning; \textbf{ReMoDetect}~\cite{lee2024remodetect}, which improves detection performance by enhancing the reward model’s ability;  and \textbf{ImBD}~\cite{chen2025imitate}, which aligns a scoring model with machine-like writing styles via Style Preference Optimization (SPO) and detects AI-generated text using a Style-Conditional Probability Curvature. These methods vary in architecture, training objective, and the type of synthetic data used for supervision, but all are fine-tuned or trained on human–AI text pairs generated by LLMs. Full descriptions and details are provided in Appendix~\ref{appendix:detectors_supervised_detectors}.

\subsubsection{LLM Detectors}
We evaluate eight large language models as detectors, each prompted to perform binary classification—distinguishing human-written from AI-generated text—using a standardized instruction. The full list of evaluated models, along with the exact prompt template and inference settings, is provided in Appendix~\ref{appendix:detectors_llm_detectors}.

\subsection{Metrics}
We report two primary evaluation metrics:  
\textbf{Accuracy (Acc)} and \textbf{Area Under the ROC Curve (AUROC)}.  
Accuracy measures the proportion of correctly classified samples (human-written vs. AI-generated) under a fixed decision threshold.  
AUROC evaluates the detector’s ranking performance across all possible thresholds, providing a threshold-invariant assessment that is particularly informative for imbalanced or domain-shifted settings.  

\section{Evaluation Protocol}
We evaluate AI-generated text detectors across seven settings within and beyond C-ReD: in-distribution performance, generalization to unseen generators, cross-domain transfer (with fixed generators), prompt complexity effects, LLM-as-detector reliability under multiple prompting strategies, and out-of-distribution validation on external Chinese datasets. For each domain and each source, we pre-split the data into fixed training and test subsets before any experiments.

\subsection{Domain- and Generator-Agnostic Evaluation on C-ReD}
A major requirement for practical AI-generated text detectors is robustness across diverse writing domains and unknown generation models. To evaluate this, we assess all zero-shot and supervised methods under a unified protocol in C-ReD: at inference time, detectors perform binary classification—distinguishing human-written from AI-generated text—without access to generator identity or domain labels. 
The test set covers all five domains, each containing AI-generated texts from all nine LLMs paired with human-written counterparts. 
Results are reported using the AUROC metric. No method-specific tuning or adaptation is applied: zero-shot detectors use fixed reference models, and supervised detectors rely solely on publicly released pre-trained weights without fine-tuning.

\subsection{Training on C-ReD: In-Distribution and Out-of-Distribution Evaluation}
A key challenge in AI-generated text detection is building models that generalize beyond the specific generators seen during training. To address this, we fine-tune OpenAI Detector and ImBD on a diverse, domain-balanced training set from seven LLMs in C-ReD, except Claude-3.5-Haiku and Gemini-2.5-Flash. The resulting models are evaluated on test data from all nine LLMs—including these two held-out models—using the AUROC metric to assess both in-distribution performance and robustness to previously unseen generators. This setup directly probes whether training on C-ReD’s multi-generator, multi-domain data can improve both reliability and transferability of supervised detectors—a crucial step toward real-world deployment.

\subsection{Domain Generalization Under Fixed Generators}
An important challenge in AI-generated text detection is whether models trained on one writing domain can generalize to others. To isolate the effect of domain shift from generator variability, we evaluate cross-domain generalization under controlled conditions: using only two representative LLMs (Qwen2.5 and GPT-4o) as fixed generators throughout training and testing.
For each generator, we train a supervised detector on human–AI text pairs from one source domain and evaluate it on all five domains without access to domain labels at inference time. 


\subsection{Evaluating LLMs as AI-Text Detectors}
To assess whether large language models can reliably distinguish human-written from AI-generated text, we design a controlled detection task across the five domains in C-ReD. For each domain, we evaluate three representative generators—Qwen2.5, GPT-4o, and Deepseek-R1—against a diverse set of Judge LLMs.
We compare three prompting strategies:
(1) \texttt{normal} (zero-shot),
(2) \texttt{context} (few-shot with three human-AI example pairs), and
(3) \texttt{description} (rule-based, using a Qwen3-generated stylistic summary).
\begin{CJK*}{UTF8}{gbsn}
All prompts enforce a strict binary output format ({\footnotesize``机器生成''} or {\footnotesize``人类生成''}) for automatic evaluation. Full prompt templates are provided in Appendix~\ref{appendix:llm_detetct_prompting_strategies}.
\end{CJK*}

\subsection{Ablation on Prompt Complexity in Academic Writing}
To investigate the effect of prompt complexity on detection difficulty, we generate AI-authored text in the Academic Writing domain using both the original C-ReD prompt and a simplified variant (see Appendix~\ref{appendix:prompt_complexity_simplified_prompt_design} for design details). For each prompt type, we collect outputs from Qwen2.5 and GPT-4o and pair them with human-written texts from the same topic distribution. Detectors are then evaluated on both sets under the standard binary classification protocol using AUROC, without access to prompt information at inference time.

\subsection{Evaluation on External Traditional Chinese News}
To assess the robustness of detection methods in a distinct linguistic and stylistic setting, we evaluate all three core protocols on an external dataset of Traditional Chinese news articles. This dataset comprises human-written texts and AI-generated counterparts produced by all nine LLMs in C-ReD, each prompted natively in Traditional Chinese.
We replicate the same evaluation setups as in the main benchmark:
\begin{itemize}
    \item \textbf{Domain- and Generator-Agnostic Evaluation}: Detectors are applied without adaptation, with no access to domain or generator labels at inference time.
    \item \textbf{Training and Evaluation on Traditional Chinese Data}: Supervised detectors are fine-tuned on a domain-balanced training set constructed from the Traditional Chinese news data (covering seven of the nine LLMs), and evaluated on the full test set—including the two held-out generators.
    \item \textbf{LLM-as-Detector Evaluation}: Judge LLMs use the three prompting strategies (\texttt{normal}, \texttt{context}, \texttt{description}).
\end{itemize}

All evaluations follow the standard binary classification protocol using AUROC. This setup provides an independent assessment of detection performance in a real-world, externally collected Traditional Chinese news domain—complementing the main C-ReD and revealing how methods behave under script and stylistic distribution shifts.

\subsection{Validation via Transfer Performance on External Chinese Datasets}
To assess generalization, we fine-tune OpenAI Detector and ImBD exclusively on C-ReD’s multi-generator Q\&A training set (covering seven LLMs). We evaluate these models—before and after fine-tuning—on the Chinese QA subsets of \textbf{M4}~\cite{wang2023m4}, which contain human–AI text pairs generated by \textbf{ChatGPT} and \textbf{davinci-003}. Although both datasets are QA-oriented, they differ in data source, style, and generation models. Since the M4 Chinese samples were not used in C-ReD’s construction, this provides a strict out-of-distribution test to evaluate whether detectors trained on C-ReD’s Q\&A domain generalize to real-world Chinese QA content. Full dataset details are in Appendix~\ref{appendix:exetrnal_chinese_datasets_test_dataset}.

\section{Results and Discussion}

\subsection{Domain- and Generator-Agnostic Evaluation on C-ReD}
Detection performance varies significantly across domains and LLM generators—two critical factors often overlooked in prior work. Table~\ref{tab:main_results} shows the average AUROC across five domains, with full per-model results in Appendix~\ref{appendix:main_results}.
\textbf{Domain difficulty is highly non-uniform.} Detection is consistently easier in structured, stylistically constrained domains like Q\&A and Film Reviews, where AI-generated text exhibits clear statistical anomalies. In contrast, performance drops in News and Academic writing, where modern LLMs produce fluent, coherent prose that closely mimics human style.
\textbf{Generator characteristics further modulate detectability.} For instance, Deepseek-R1 frequently employs chain-of-thought (CoT) reasoning, yielding outputs with enhanced logical coherence and structure—properties that obscure typical AI-generation artifacts and make detection more challenging. This effect is evident in the low AUROC scores of multiple methods on Deepseek-R1. Moreover, existing detectors exhibit systemic limitations under generator shift: zero-shot methods relying on a fixed reference model suffer from architecture- or style-mismatch when applied to unseen LLMs, while supervised models like RoBERTa fail on modern generators due to training data bias—having been trained primarily on older, lower-quality AI text.

\begin{table}[t]
\centering
\small        
\setlength{\tabcolsep}{3pt}  
\renewcommand{\arraystretch}{0.95} 
\caption{
AUROC performance across five domains. Results are averaged over 9 LLMs.
}
\label{tab:main_results}
\begin{tabular}{@{}l c c c c c@{}}  
\toprule
\textbf{Method} & \textbf{Film} & \textbf{Comp.} & \textbf{Q\&A} & \textbf{News} & \textbf{Acad.} \\
\midrule
Log-Likelihood     & 0.8344 & 0.8433 & 0.9343 & 0.7373 & 0.7326 \\
Entropy            & 0.8427 & 0.8088 & 0.9253 & 0.6873 & 0.6907 \\
Log-Rank            & 0.8286 & 0.8473 & 0.9301 & 0.7372 & 0.7384 \\
LRR                & 0.7133 & 0.8473 & 0.8388 & 0.7004 & 0.7231 \\
Fast-DetectGPT     & 0.6999 & 0.8952 & 0.8385 & 0.7626 & 0.7132 \\
Lastde             & 0.5668 & 0.6500 & 0.6403 & 0.6762 & 0.7468 \\
Lastde++           & 0.7059 & 0.8821 & 0.8148 & 0.7905 & 0.7157 \\
DNA-DetectLLM      & 0.7595 & 0.9263 & 0.8439 & 0.7231 & 0.6226 \\
LAPD      & 0.8857 & 0.9528 & 0.9726 & 0.9407 & 0.9150 \\
\midrule
RoBERTa-base & 0.6461 & 0.5191 & 0.5139 & 0.4937 & 0.4316 \\
RoBERTa-large & 0.6121 & 0.5393 & 0.5255 & 0.3699 & 0.4059 \\
RADAR        & 0.8291 & 0.6338 & 0.7605 & 0.4638 & 0.5167 \\
ReMoDetect & 0.9731 & 0.8731 & 0.9755 & 0.8652 & 0.9126 \\
ImBD               & 0.8760 & 0.9140 & 0.9011 & 0.7953 & 0.8056 \\
\bottomrule
\end{tabular}
\end{table}

\subsection{Training on C-ReD: In-Distribution and Out-of-Distribution Evaluation}
We analyze detection performance under two evaluation settings: pre-trained and fine-tuning on C-ReD. The test set includes nine LLMs, seven of which are part of the C-ReD training distribution (in-distribution, ID), while the remaining two—Claude-3.5-Haiku and Gemini-2.5-Flash—are held out (out-of-distribution, OOD). Full results are reported in Appendix~\ref{appendix:finetune_comparison}.
Fine-tuning on C-ReD leads to a dramatic improvement across all domains, confirming that supervised detectors heavily rely on domain- and generator-aligned training data. Importantly, this gain extends consistently to OOD generators, demonstrating that \textbf{C-ReD’s diverse composition enables meaningful generalization beyond the specific models seen during training}. Nevertheless, a small gap between ID and OOD performance remains after fine-tuning, indicating that while C-ReD greatly reduces generator bias, perfect transfer to new commercial models is still challenging. Notably, Deepseek-R1—the hardest ID model to detect—owes its low detectability to its chain-of-thought style, which generates highly coherent, human-like reasoning. This underscores that intrinsic generation traits, not just model identity, fundamentally limit detection reliability.

\subsection{Cross-Domain Generalization Analysis Under Fixed Generators}
\begin{figure}[t]
    \centering
    \begin{subfigure}{1\linewidth}
        \centering
        \includegraphics[width=\linewidth]{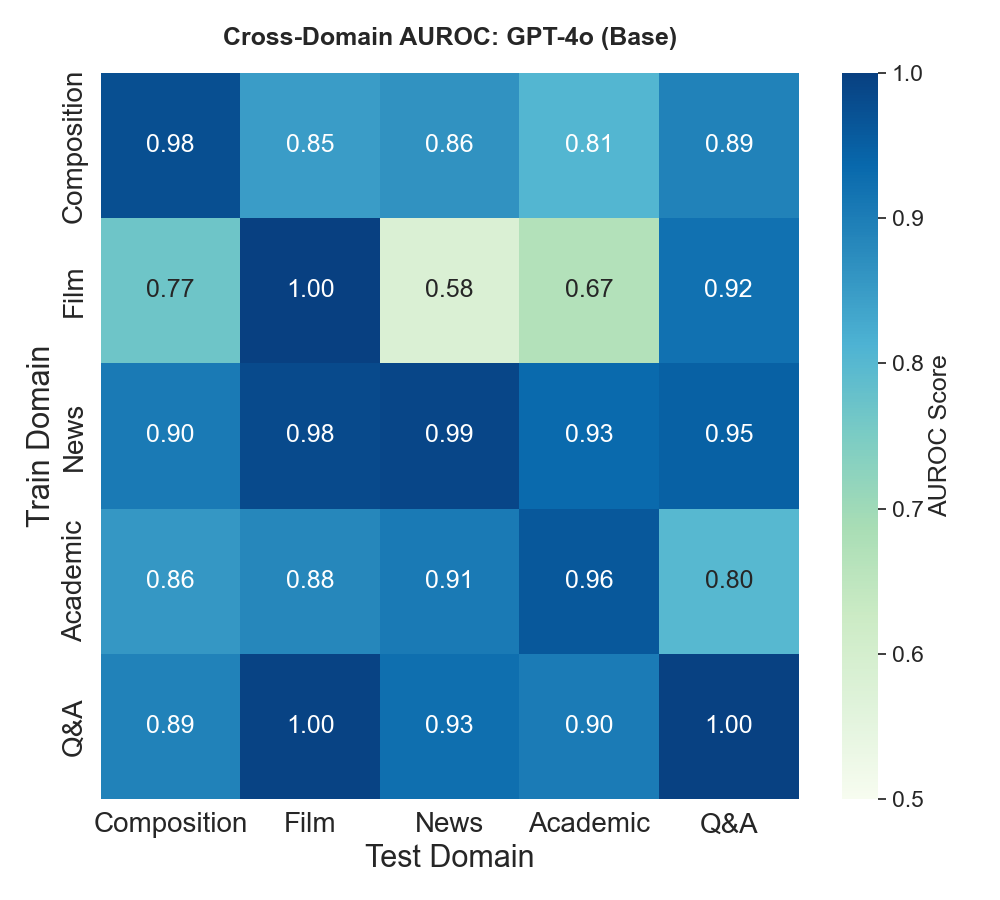}
        \caption{GPT-4o}
        \label{fig:gpt4o_base}
    \end{subfigure}
    \\
    \begin{subfigure}{1\linewidth}
        \centering
        \includegraphics[width=\linewidth]{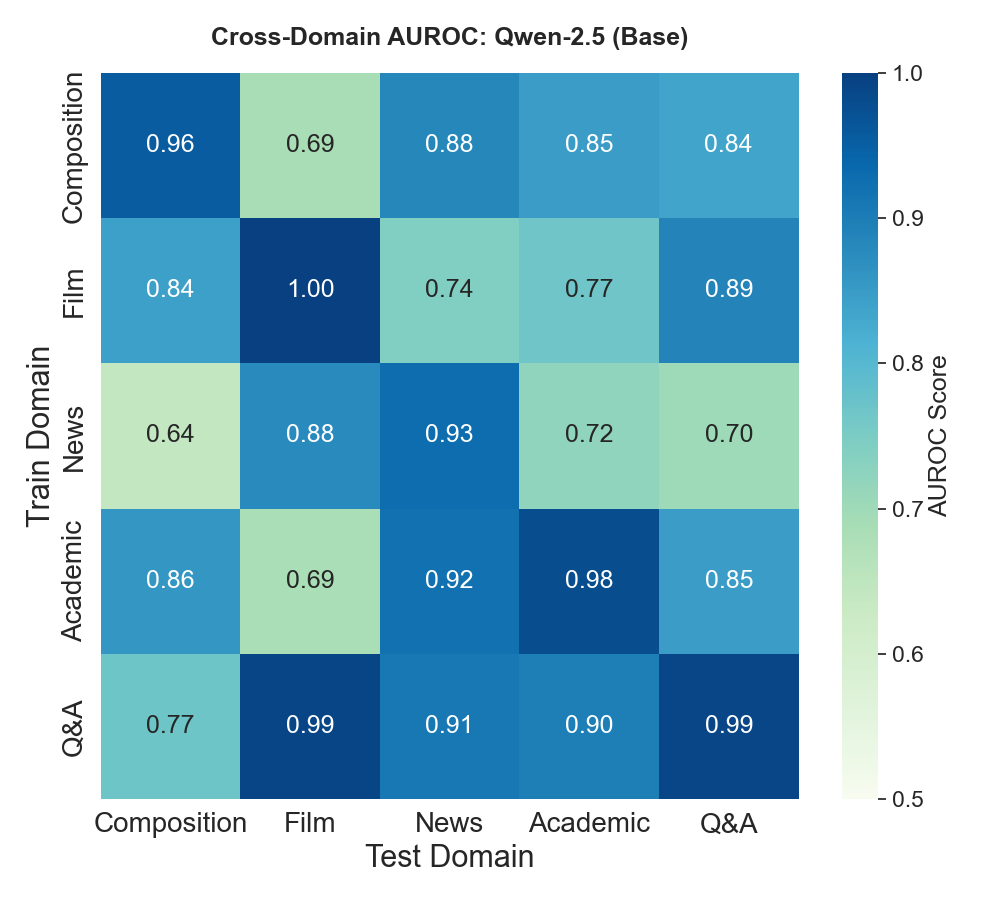}
        \caption{Qwen2.5}
        \label{fig:gpt4o_large}
    \end{subfigure}
    
    \caption{Cross-Domain AUROC Heatmaps on RoBERTa-base Model.}
    \label{fig:cross_domain_heatmaps}
\end{figure}

We examine how detection performance varies when models are trained on one domain and evaluated across all five domains. As shown in Figure~\ref{fig:gpt4o_base} and ~\ref{fig:gpt4o_large}, detectors achieve strongest performance on their training domain, but generalization to other domains is highly uneven—depending on both the source training domain and the target test domain. \textbf{Training on fluent, information-dense domains like Q\&A often leads to stronger cross-domain generalization} than training on highly stylized or opinion-driven domains like Film Review. Full numerical results for all model–domain combinations are provided in Appendix~\ref{appendix:domain_generalization_fixed_gen}.

\subsection{Analyzing LLMs as AI-Text Detectors}
Our evaluation reveals that LLMs generally fail at zero-shot detection (\texttt{normal})—even on text generated by themselves—but achieve substantially higher accuracy when given domain-specific context (\texttt{context}) or stylistic descriptions (\texttt{description}). This indicates that LLMs’ detection performance can be substantially improved through the incorporation of external contextual or stylistic cues—capabilities absent in zero-shot scenarios. Performance gains are consistent across domains and models, though challenges remain in highly structured domains like academic writing. Full results are in Appendix~\ref{appendix:llm_detect_full_results}.

\subsection{Ablation on Prompt Complexity in Academic Writing}
Our study indicates that AI-generated texts using simplified prompts are nearly as hard to detect as those created with more complex C-ReD prompts. This is consistent for both Qwen2.5 and GPT-4o, showing that these models can produce human-like academic text even with reduced instructions. The minimal difference in detection performance suggests that strong LLMs maintain stylistic and structural coherence regardless of prompt complexity, making their outputs challenging to distinguish. Full results are reported in Table~\ref{tab:prompt_complexity_ablation} and results of bootstrap-based statistical analysis are reported in Appendix~\ref{appendix:prompt_complexity_simplified_prompt_design}.

\subsection{Evaluation on External Traditional Chinese News}

We evaluate all three core protocols on an external Traditional Chinese news dataset to assess detection robustness under a different linguistic and stylistic setting. The results closely mirror those observed in the main Simplified Chinese benchmark: supervised models achieve strong in-domain performance but still show a slight drop on held-out generators; LLM-as-detector approaches remain highly sensitive to prompting strategy and generator style. This consistency across writing systems suggests that key detection challenges—such as the impact of domain structure and reasoning-intensive generation—are not language-specific but reflect broader, cross-lingual patterns. Full results are provided in Appendix~\ref{appendix:tc_results}.

\begin{table}[t]
\centering
\footnotesize 
\setlength{\tabcolsep}{3pt} 
\caption{AUROC scores for detecting AI-generated academic writing under original vs. simplified prompts, before and after fine-tuning on C-ReD.}
\label{tab:prompt_complexity_ablation}
\begin{tabular}{lcccc}
\toprule
\multirow{2}{*}{Model} & \multirow{2}{*}{Prompt} & \multirow{2}{*}{Generator} & \multicolumn{2}{c}{AUROC} \\
\cmidrule(lr){4-5}
& & & Baseline & Finetuned \\
\midrule
RoBERTa-base & Original    & GPT-4o   & 0.5010 & 0.9566 \\
             & Simplified  & GPT-4o   & 0.4987 & 0.9201 \\
             & Original    & Qwen2.5  & 0.3313 & 0.9648 \\
             & Simplified  & Qwen2.5  & 0.3272 & 0.9501 \\
\midrule
RoBERTa-large& Original    & GPT-4o   & 0.4032 & 0.9868 \\
             & Simplified  & GPT-4o   & 0.3937 & 0.9691 \\
             & Original    & Qwen2.5  & 0.3685 & 0.9906 \\
             & Simplified  & Qwen2.5  & 0.3954 & 0.9829 \\
\bottomrule
\end{tabular}
\end{table}

\subsection{Validation via Transfer Performance on External Datasets with Chinese Content}
As shown in Table~\ref{tab:m4_compact_transfer_performance}, detectors trained solely on C-ReD’s Q\&A domain exhibit strong transferability to external Chinese datasets. Both OpenAI Detector and ImBD show substantial improvements after fine-tuning, demonstrating that C-ReD provides effective supervision even for out-of-distribution generators. Performance on davinci-003 samples is lower but still improves significantly, suggesting greater stylistic divergence from C-ReD’s training distribution. These results confirm that fine-tuning on C-ReD enables robust generalization to real-world Chinese AI-text detection scenarios.

\begin{table}[t]
\centering
\footnotesize
\caption{
Transfer performance on \textbf{M4}~\cite{wang2023m4}. 
Detectors are trained only on C-ReD’s Q\&A domain (7 LLMs). 
Results report AUROC score.
}
\label{tab:m4_compact_transfer_performance}
\begin{tabular}{l c c c c}
\toprule
\multicolumn{1}{c}{\textbf{Detector}} & \multicolumn{2}{c}{\textbf{ChatGPT}} & \multicolumn{2}{c}{\textbf{davinci-003}} \\
\cmidrule(lr){2-3} \cmidrule(lr){4-5}
& Pre & Post & Pre & Post \\
\midrule
Roberta-base & 0.6055 & \textbf{0.8169} & 0.6354 & \textbf{0.6466} \\
Robeata-lagre & 0.5684 & \textbf{0.8890} & 0.3990 & \textbf{0.7445} \\
ImBD & 0.9751 & \textbf{0.9918} & 0.9756 & \textbf{0.9818} \\
\bottomrule
\end{tabular}
\end{table}

\section{Conclusion}
In this work, we present \textbf{C-ReD}: a comprehensive \textbf{C}hinese \textbf{Re}al-prompt AI-generated text \textbf{D}etection benchmark. Spanning five domains and nine LLMs—including leading domestic Chinese models—C-ReD enables rigorous evaluation of detection methods across in-domain performance, cross-domain generalization, out-of-distribution robustness, and cross-lingual transfer. Our extensive experiments reveal that detection difficulty is strongly influenced by both domain structure and generator characteristics, with reasoning-intensive models like Deepseek-R1 posing the greatest challenge. Crucially, fine-tuning on C-ReD yields substantial gains not only on seen generators but also on unseen commercial models and external datasets, demonstrating its representativeness and practical utility. We hope C-ReD will serve as a reliable foundation for future research and deployment of AI-generated text detectors in Chinese contexts.

\section*{Limitations}

Despite its breadth, C-ReD has several limitations. First, it focuses exclusively on Chinese text; while our preliminary results on Traditional Chinese suggest that key detection challenges generalize across writing systems, broader multilingual coverage remains future work. Second, although C-ReD includes nine LLMs, the rapid pace of model development means that new architectures or reasoning paradigms may emerge that are not represented in the current release. Third, our prompt designs, while inspired by real-world use cases, cannot capture the full spectrum of user behaviors—particularly adversarial or highly customized prompting strategies aimed at evading detection. Finally, our selection of human-written reference texts is limited in scope, and may not fully represent the diversity of writing styles, registers, or domains present in real-world Chinese content.
\section*{Acknowledgments}
This work is supported in part by the National Natural Science Foundation of China under grant 62301189, 62576122, 62571298, Guangdong Basic and Applied Basic Research Foundation under grant 2026A1515011139.


\bibliography{custom}

\clearpage
\appendix

\section{Dataset Construction}
\subsection{Data Source}
\label{appendix:dataset_construction data_source}
\noindent\textbf{News} \ \ 
We selected 3,000 news articles from THUCNews~\cite{li2006comparison}, a dataset derived from Sina News RSS feeds (2005–2011). The sample is balanced across five categories—sports, politics, finance, entertainment, and education—with 600 articles per category. To ensure suitability for Chinese-language detection, we filtered out articles with high English content, removed metadata (e.g., headlines, publisher information), and merged each article into a single paragraph. All texts were further constrained to a maximum length of 800 words.

\noindent\textbf{Q\&A} \ \ 
We sourced 2,956 human-written answers from Zhihu-KOL~\cite{zhihukol}, a dataset collected from Zhihu—a major Chinese Q\&A and knowledge-sharing platform. The sample covers 2,892 distinct questions and includes answers with lengths uniformly distributed between 100 and 850 words, ensuring diversity in response length. To enhance data quality and language consistency, we filtered out responses with high English content, repaired formatting issues (e.g., irregular spacing and missing punctuation), and merged each answer into a single coherent paragraph.

\noindent\textbf{Film Review} \ \
We focus on film reviews as a representative short-text scenario where LLMs are widely applied. We collected 2,960 user reviews from Douban—a leading Chinese film review platform—sourced from the ChineseNlpCorpus~\cite{AesopChowChineseNLPCorpus}, which contains reviews for 28 movies. To ensure consistency and suitability for detection tasks, we selected reviews with lengths between 100 and 200 words. We then cleaned the data by removing excessive line breaks, restoring missing punctuation, and merging each review into a single paragraph.

\noindent\textbf{Composition} \ \
We include Gaokao model essays as a representative genre of Chinese writing frequently generated by LLMs. We collected 1,081 high-scoring essays from China’s National College Entrance Examination by crawling publicly available websites. Given the heterogeneous sources, we standardized the data by removing titles, author names, source URLs, editorial comments, and other metadata, retaining only the main essay body. We also cleaned formatting artifacts such as tab characters and excessive whitespace to ensure textual consistency, resulting in a clean and uniform dataset.

\noindent\textbf{Academic Writing} \ \
To cover the academic writing domain, we extracted 500 Chinese-language research papers from ChinaXiv~\cite{chinaxiv}, an open repository for Chinese scientific literature. From each paper (originally in PDF format), we parsed the title, keyword, abstract, and introduction sections. We removed non-textual elements such as figures, tables, and citations, and further cleaned the extracted text by fixing irregular whitespace, spurious line breaks, and other formatting artifacts, resulting in a clean and standardized academic text dataset.

\noindent\textbf{TC News} \ \
In addition to Simplified Chinese, we also include Traditional Chinese content by selecting 2,000 news articles from News-Collection-Zhtw~\cite{news-collection-zhtw}, covering four categories: article, tech, science, and daily-weekly. To ensure balance, we sampled 500 articles per category. We applied the same cleaning procedure as used for the Simplified Chinese news dataset—removing metadata, filtering out high-English-content articles, and merging text into single paragraphs—and further constrained article lengths to 100–600 words.

\subsection{Generative LLMs}
\label{appendix:dataset_construction generative_llms}
All AI-generated texts in this work were produced by querying APIs of nine large language models. This black-box setup closely mirrors real-world usage, where end users interact with LLMs solely through cloud-based interfaces without access to internal weights or decoding details. To ensure a fair and controlled comparison across models, we applied an identical generation protocol for each data type. Same prompt templates were used for all models within a given domain and we fixed decoding parameters where supported by the API. Below is a brief overview of the nine models included in our study:

\noindent\textbf{OpenAI} (\texttt{gpt-3.5-turbo}, \texttt{gpt-4o}): Proprietary models with strong multilingual capabilities, including fluent Chinese generation.

\noindent\textbf{Google} (\texttt{Gemini-2.5-Flash}): A fast, lightweight model optimized for low-latency applications while maintaining decent Chinese text quality.

\noindent\textbf{Anthropic} (\texttt{Claude-3.5-Haiku}): The fastest model in the Claude 3.5 series, designed for instant responses and robust instruction following in Chinese.

\noindent\textbf{Deepseek} (\texttt{Deepseek-V3}, \texttt{Deepseek-R1}): Open-weight bilingual models; Deepseek-R1 is enhanced for reasoning via chain-of-thought prompting.

\noindent\textbf{Qwen} (\texttt{Qwen2.5}, \texttt{Qwen3}): State-of-the-art open-source models from Alibaba, excelling in Chinese comprehension and generation.

\noindent\textbf{Doubao} (\texttt{Doubao-1.5-Pro}): ByteDance’s proprietary assistant model, fine-tuned for Chinese writing tasks such as exam essays and creative composition.

\subsection{Prompt Design and Examples}
\label{appendix:dataset_construction prompt_design_and_examples}
To make the AI-generated topics comparable to human-written text, we first analyzed the human corpus for each domain (e.g., examining all movies covered in the review dataset, or breaking down news by category), then used an LLM to generate multiple candidate prompts for each domain, and manually screened them through iterative refinement to select the highest-quality ones.
\tcbset{
  promptbox/.style={
    enhanced,
    colback=gray!3,        
    colframe=gray!30,      
    boxrule=0.4pt,
    arc=2pt,               
    left=6pt, right=6pt, top=4pt, bottom=4pt,
    fontupper=\small\ttfamily, 
    boxsep=0pt,
    before skip=6pt,
    after skip=6pt
  }
}

\subsubsection{Composition}
Our prompt design is based on official Gaokao composition questions, covering three common formats. We adopt the standard instruction phrasing from recent exams, parameterized by the composition title and target word count. A representative example is shown below:

\begin{CJK*}{UTF8}{gbsn}
\begin{tcolorbox}[promptbox]
请以\{composition title\}为题，写一篇作文。要求：自选角度，自行立意；除诗歌外，文体不限；字数控制在\{word count\}左右。
\end{tcolorbox}
\end{CJK*}

The prompt structure closely follows official Gaokao guidelines, enabling LLMs to generate essays in a manner consistent with how users would interact with them for composition practice in real-world educational contexts.

\subsubsection{Film Review}
We leveraged LLMs to develop a set of structured review guidelines tailored to common film genres in our dataset. From these, we selected 10 representative templates that capture diverse critical perspectives (e.g., emotional response, thematic analysis). Each prompt is parameterized by the film title and target word count to guide LLM generation. A representative example is shown below:

\begin{CJK*}{UTF8}{gbsn}
\begin{tcolorbox}[promptbox]
以第一人称视角描述观看\{film title\}的体验，包括最触动你的1-2个场景及原因以及电影如何引发你的思考。要求：字数为\{word count\}字左右。
\end{tcolorbox}
\end{CJK*}

This design mirrors real-world practices where AI-generated reviews are typically produced by specifying film-specific instructions—such as perspective, focus, and length—to steer LLM output, thereby enhancing relevance and personalization.

\subsubsection{Q\&A}
We developed structured answer templates using LLMs, informed by an analysis of question-answer patterns in our dataset. From this, we selected 10 representative templates that support diverse response styles. Each prompt combines a specific question and target word count with a standardized instruction to guide coherent and contextually appropriate answers. A representative example is shown below:

\begin{CJK*}{UTF8}{gbsn}
\begin{tcolorbox}[promptbox]
你是一位资深的网络话题评论员，目前你在网络上遇到了以下话题：\{question\}，请从一位用户的角度来对此话题进行分析以及探讨，要求内容字数在\{word count\}左右。
\end{tcolorbox}
\end{CJK*}

This approach reflects a common real-world practice in which LLMs are guided by both the question content and explicit formatting constraints to generate user-like responses—enabling us to simulate authentic human-AI interaction scenarios in online discussion contexts.

\subsubsection{News}
We developed a total of 25 structured news writing templates—five for each of five major news categories—using LLMs. These templates were derived through iterative analysis of real articles in our dataset. This ensures that LLM-generated content adheres not only to factual context but also to genre-specific stylistic norms. Our instruction-based approach better reflects real-world AI-assisted journalism workflows, where LLMs are typically provided with explicit guidelines rather than asked to continue from a headline alone. A representative example from the sports category is shown below:

\begin{CJK*}{UTF8}{gbsn}
\begin{tcolorbox}[promptbox]
请以体育新闻报道的风格，撰写一篇关于\{title\}的文章。内容要包括比赛过程、运动员表现和专业分析，语言要生动有激情，使用体育专业术语，字数控制在\{word count\}左右。
\end{tcolorbox}
\end{CJK*}

\subsubsection{Academic Writing}
We implement two distinct generation modes to simulate different stages of academic writing assistance:

\paragraph{Abstract Drafting from Title and Keywords}
In early-stage ideation, researchers often draft abstracts based on a working title and preliminary keywords. To reflect this scenario, we construct prompts using only the paper title and keywords, instructing the LLM to generate a concise abstract that outlines the research background, core problem, methodology, key findings, and significance.

\begin{CJK*}{UTF8}{gbsn}
\begin{tcolorbox}[promptbox]
请根据以下信息撰写一篇通用型学术论文摘要。
\\ 文章标题：\{title\}
\\ 文章关键词：\{keyword\}
\\ 摘要应简明扼要地说明研究背景、核心问题、研究方法、主要发现（或论点）及研究意义。要求语言严谨、逻辑连贯、术语准确，字数控制在\{word count\}字左右，适合投稿至各类中文学术期刊。
\end{tcolorbox}
\end{CJK*}

\paragraph{Introduction Generation from Abstract}
Once an abstract is available, researchers typically expand it into a full introduction. We model this common workflow by providing the LLM with the title, keywords, and abstract to generate an introduction that frames the scientific question and motivates the study.

\begin{CJK*}{UTF8}{gbsn}
\begin{tcolorbox}[promptbox]
请以提出关键科学问题的方式撰写引言内容，引导读者思考研究价值。
\\ 文章标题：\{title\}
\\ 文章关键词：\{keyword\}
\\ 文章摘要：\{abstract\}
\\ 字数控制在\{word count\}字左右，语言流畅，兼具学术性与可读性。
\end{tcolorbox}
\end{CJK*}

This dual-mode design captures realistic AI-assisted writing practices across different phases of scholarly composition.

\subsubsection{TC News}
We applied the same template-based approach as in simplified Chinese news to traditional Chinese (TC) journalism, developing 20 structured prompts across four categories: general news, technology, science, and weekly digests (5 templates per category). Each prompt combines a headline and target word count with genre-specific instructions to guide LLMs in producing authentic TC journalistic content.
An example from the weekly digest category:

\begin{CJK*}{UTF8}{bsmi}
\begin{tcolorbox}[promptbox]
根據標題\{title\}，製作一份每週要點回顧。要求以分項列舉的形式（如‘重點事件’、‘數據解讀’、‘觀點彙總’等）清晰呈現一週的核心信息，語言需精煉、準確，字數控制在\{word count\}左右。
\end{tcolorbox}
\end{CJK*}

\subsection{AI Generation Constraints}
\label{appendix:dataset_construction ai_generation_constrains}
To ensure that AI-generated texts faithfully reflect the characteristics of human-written counterparts in each domain, we derived key generation constraints directly from the full empirical distribution of the human data. Specifically:

\begin{itemize}
    \item \textbf{Length (words)}: The min–max range for each domain was set to cover 100\% of the human-written samples’ lengths, allowing slight extrapolation while preserving genre-typical scale.
    \item \textbf{Minimum Chinese Ratio}: Defined as the percentage of Chinese characters among all characters, this threshold filters out generations with excessive English or symbolic noise, aligning with the linguistic purity of cleaned human texts.
    \item \textbf{Maximum Repetition Length}: The maximum allowed length (in characters) of any repeated substring within the generated text. This threshold is set based on the longest duplicated span observed in human-written samples, with a small safety margin; generations containing longer repeated sequences are typically indicative of model degeneration or redundant hallucination.
\end{itemize}

These thresholds were enforced during both real-time generation monitoring and post-generation filtering. As a result, the AI-generated corpus maintains strong distributional alignment with human-written texts across structural and linguistic dimensions. The exact parameter values for each domain are summarized in Table~\ref{tab:ai_generation_constraints}.

\begin{table}[t]
\centering
\footnotesize
\caption{Quality control thresholds for AI-generated texts by domain.}
\label{tab:ai_generation_constraints}
\begin{tabular}{@{}lccc@{}}
\toprule
Domain & Length & Min. Ch & Max. Rep \\
       & (words) & Ratio (\%) & (chars) \\
\midrule
News            & 80–1100   & 80   & 20 \\
Q\&A            & 80–1000   & 80   & 25 \\
Film Review     & 80–300   & 80   & 15 \\
Composition     & 100–2000   & 80   & 25 \\
Academic Writing  & 80–1600   & 60   & 40 \\
TC News   & 80–900   & 80   & 20 \\
\bottomrule
\end{tabular}

\end{table}

\subsection{Text Length Statistics}
\label{appendix:text_length_statistics}
We observed that different large language models exhibit varying length biases when generating text—some tend to produce longer outputs than human-written text, while others generate shorter ones, indicating that text length alone cannot reliably distinguish between AI-generated and human-written content. Full details are showed in Table~\ref{tab:length_stats}.

\subsection{CSV Schema}
\label{appendix:dataset_construction_csv_shcema}
While Table~\ref{tab:csv_schema} defines the core fields shared across all samples, each domain includes additional metadata relevant to its source or generation context. These domain-specific fields are listed below. \newline
\noindent\textbf{Film Review} \\
\texttt{film\_title} (str) — Title of the reviewed film.

\noindent\textbf{Composition} \\
\texttt{composition\_title} (str) — Title of composition. \\
\texttt{year} (int) — Year the composition was written.

\noindent\textbf{News} \\
\texttt{news\_title} (str) — Headline of the news article. \\
\texttt{news\_category} (str) — Editorial category.

\noindent\textbf{Q\&A} \\
\texttt{question\_title} (str) — The original question.

\noindent\textbf{Academic Writing} \\
\texttt{paper\_title} (str) — Title of the source paper. \\
\texttt{keyword} (str) — Comma-separated keywords. \\
\texttt{category} (str) — \texttt{abstract} or \texttt{introduction}. \\
\texttt{extra\_text} (str) — Contextual supplement: empty string (\texttt{""}) for abstracts; contains the full abstract text when the sample is an introduction.

All domain-specific fields are stored as string or integer types and appear only in samples belonging to the respective domain. They are absent in other domains. 

\subsection{Statistics}
\label{appendix:dataset_construction statistics}
Our dataset consists of 128,610 texts across five domains, with contributions from ten text sources (including one human reference and nine large language models). The complete distribution is shown in Table~\ref{tab:dataset_distribution}. 

 All models generate roughly the same number of samples as the human reference within each domain, ensuring a fair comparison. Minor discrepancies arise from post-generation filtering, particularly for models like \textit{Gemini-2.5-Flash}, which shows slightly fewer samples in certain domains.

\renewcommand{\theadfont}{\bfseries}
\renewcommand{\theadalign}{c} 

\begin{table*}[t]
\centering
\fontsize{8pt}{10pt}\selectfont   
\setlength{\tabcolsep}{4pt}       
\begin{tabular}{@{}l*{11}{c}@{}}
\toprule
\textbf{Domain} &
\textbf{Human} &
\thead{GPT-3.5- \\ Turbo} &
\textbf{GPT-4o} &
\thead{Deepseek- \\ V3} &
\thead{Deepseek- \\ R1} &
\textbf{Qwen2.5} &
\textbf{Qwen3} &
\thead{Doubao- \\ 1.5-pro} &
\thead{Gemini- \\ 2.5-Flash} &
\thead{Claude- \\ 3.5-Haiku} &
\textbf{Total} \\
\midrule
Composition & 1,081 & 1,079 & 1,073 & 1,080 & 1,081 & 1,081 & 1,077 & 1,078 & 1,078 & 1,081 & 10,789 \\
Film Review & 2,960 & 2,959 & 2,960 & 2,960 & 2,960 & 2,960 & 2,960 & 2,960 & 2,960 & 2,960 & 29,599 \\
Q\&A        & 2,956 & 2,933 & 2,864 & 2,943 & 2,869 & 2,952 & 2,942 & 2,939 & 2,719 & 2,954 & 29,071 \\
News        & 3,000 & 2,997 & 2,988 & 2,999& 2,969 & 2,995 & 2,999 & 2,966 & 2,846 & 2,999 & 29,758 \\
Academic Writing    & 1,000 & 991 & 1,000 & 1,000 & 999 & 1,000 & 1,000 & 999 & 998 & 1,000 & 9,987 \\
TC News & 2,000 & 1,998 & 2,000 & 1,996 & 1,972 & 1,970 & 1,964 & 1,989 & 1,517 & 2,000 & 19,406 \\
\midrule
\textbf{Total}     & 12,997 & 12,957 & 12,885 & 12,978 & 12,850 & 12,958 & 12,942 & 12,931 & 12,118 & 12,994 & 128,610 \\
\bottomrule
\end{tabular}
\caption{Distribution of the dataset across domains and text sources (number of samples).}
\label{tab:dataset_distribution}
\end{table*}

\section{Detectors}
\subsection{Zero-shot Detectors}
\label{appendix:detectors_zero-shot_detectors}
Zero-shot detectors refer to methods that distinguish AI-generated text from human-written text without any task-specific training or labeled examples. Instead, they rely solely on the statistical properties of text under a pre-trained language model (often the same model used for generation, or a comparable reference model). These approaches typically compute scores based on token-level likelihoods, prediction ranks, entropy, or other intrinsic signals derived during forward passes through the model. 
Below we summarize the zero-shot detection methods evaluated on C-ReD:
\begin{itemize}
    \item \textbf{Log-Likelihood}~\cite{solaiman2019release}: Computes the average log-probability of tokens in a text under a reference language model. AI-generated text typically exhibits higher likelihood than human-written text due to exposure during training.
    \item \textbf{Entropy}~\cite{gehrmann2019gltr}: Measures the entropy of the predicted token distribution at each position. Human-written text tends to have higher entropy (more unpredictable) compared to AI output.
    \item \textbf{Log-Rank}~\cite{mitchell2023detectgpt}: Uses the rank of each ground-truth token in the model’s vocabulary prediction sorted by probability. AI-generated tokens often appear at higher ranks.
    \item \textbf{Log-Likelihood Log-Rank Ratio (LRR)}~\cite{su2023detectllm}: A zero-shot detection score that combines log-likelihood and log-rank. By taking the ratio of these two quantities, LRR better distinguishes AI-generated text—which tends to have high likelihood but even more strongly concentrated token ranks—from human-written text.
    \item \textbf{Fast-DetectGPT} \cite{bao2024fast}: An efficient alternative to DetectGPT~\cite{mitchell2023detectgpt} that introduces conditional probability curvature as its core metric and uses a faster sampling approach.
    \item \textbf{Lastde / Lastde++}~\cite{xu2025trainingfree}: A training-free detection method that treats the sequence of token probabilities generated by a language language model as a time series. By analyzing this sequence, Lastde and Lastde++ identify distinctive patterns characteristic of AI-generated text.
    \item \textbf{DNA-DetectLLM}~\cite{zhu2025dna}: A zero-shot method that constructs an ideal AI-generated sequence and measures how much an input text must be “repaired” to match it. The repair effort—higher for human text—serves as the detection score.
    \item \textbf{LAPD}~\cite{wu2026alignment}: A zero-shot method that formalizes the alignment imprint as a measurable statistic. It introduces the Log-likelihood Alignment Preference Discrepancy, a standardized information-weighted metric that leverages the statistical signatures left by model alignment to achieve superior performance.
    
\end{itemize}
For all methods except DNA-DetectLLM and LAPD, we use \texttt{EleutherAI/gpt-j-6b} as the scoring model to ensure a consistent basis for comparison. Specifically, Fast-DetectGPT and Lastde++ also adopt GPT-J-6B as their reference model. For DNA-DetectLLM, following its dual-model design, we set the observer model to \texttt{tiiuae/falcon-7b} and the performer model to \texttt{tiiuae/falcon-7b-instruct}, as recommended in the original implementation. Also for LAPD, we use \texttt{meta-llama/Llama-2-7b} as recommended following the dual-modal design.

\subsection{Supervised Detectors}
\label{appendix:detectors_supervised_detectors}
Supervised detectors are data-driven approaches that learn to discriminate between human-written and AI-generated text by training on labeled datasets containing examples from both sources. The performance of supervised detectors heavily depends on the quality, diversity, and recency of the training data. Below we provide brief technical descriptions of the methods evaluated on C-ReD:

\begin{itemize}
    \item \textbf{OpenAI Detector}~\cite{solaiman2019release}: A detection classifier based on the pretrained RoBERTa model \cite{liu2019roberta}, originally trained to detect GPT-2 outputs. It takes raw text as input and predicts the probability of it being AI-generated. Despite being trained on older models, it remains a widely used baseline.
    \item \textbf{RADAR}~\cite{hu2023radar}: It trains a robust AI-generated text detector through adversarial learning between two language models: a paraphraser that rewrites AI-generated text to evade detection, and a detector that aims to correctly identify such rewritten outputs. By casting detection as a two-player game, RADAR enhances the detector’s generalization to adaptive or obfuscated AI-generated text without requiring access to the original generator.
    \item \textbf{ReMoDetect}~\cite{lee2024remodetect}: A detection framework that enhances a reward model’s ability by continually fine-tuning the reward model with experience replay to prevent overfitting, while also incorporating a mixed-preference dataset—created by partially rephrasing human-written text with an LLM—as an intermediate signal to refine the decision boundary.
    \item \textbf{Imitate Before Detect (ImBD)} \cite{chen2025imitate}: It first employs Style Preference Optimization (SPO) to align a scoring model with machine-like writing styles, then uses a Style-Conditional Probability Curvature (Style-CPC) metric for detection.
\end{itemize}
All supervised detectors are either used in their officially released form or re-implemented following the original papers, and fine-tuned on human–AI text pairs from the C-ReD training split. For RADAR and ReMoDetect, we utilized pre-trained models directly from HuggingFace without fine-tuning. For the OpenAI Detector and ImBD, we performed fine-tuning using specific hyperparameters. All models were tested with a maximum sequence length of 512. Detailed implementation settings and parameter configurations are listed in Table \ref{tab:impl_details}.

\begin{table*}[htbp]
    \centering
    \caption{Implementation Details and Hyperparameter Settings for Supervised Detectors}
    \label{tab:impl_details}
    
    \small              
    \renewcommand{\arraystretch}{1.2} 
    \setlength{\tabcolsep}{8pt}       
    
    \begin{tabular}{@{} l p{4cm} p{8cm} @{}}
        \toprule
        \textbf{Method} & \textbf{Model / Base} & \textbf{Configuration / Hyperparameters} \\
        \midrule
        
        \multirow{2}{*}{\textbf{RADAR}} 
        & \multirow{2}{=}{TrustSafeAI/ RADAR-Vicuna-7B} 
        & \textbf{Mode:} Pre-trained (No fine-tuning) \newline
           \textbf{Inference:} padding=True, truncation=True, max\_length=512 \\
        \midrule
        
        \multirow{2}{*}{\textbf{ReMoDetect}} 
        & \multirow{2}{=}{hyunseoki/ ReMoDetect-deberta} 
        & \textbf{Mode:} Pre-trained (No fine-tuning) \newline
           \textbf{Inference:} padding=True, truncation=True, max\_length=512 \\
        \midrule
        
        \multirow{2}{*}{\textbf{OpenAI Detector}} 
        & \multirow{2}{=}{openai-community/ roberta-base/large} 
        & \textbf{Inference:} padding=True, truncation=True, max\_length=512 \newline
           \textbf{Fine-tuning:} lr=2e-5, epochs=5, warmup=100, weight\_decay=0.01 \\
        \midrule
        
        \textbf{ImBD} 
        & EleutherAI/ gpt-j-6B 
        & \textbf{PEFT Type:} LORA (r=8, alpha=32, dropout=0.1) \newline
          \textbf{Target Modules:} ["q\_proj", "v\_proj"] \newline
          \textbf{Task:} CAUSAL\_LM \\
          
        \bottomrule
    \end{tabular}
\end{table*}

\subsection{LLM Detectors}
\label{appendix:detectors_llm_detectors}
We evaluate the following eight large language models as detectors:
\textbf{Qwen2.5}~\cite{qwen2.5},
\textbf{Qwen3}~\cite{qwen3},
\textbf{Deepseek-V3}~\cite{liu2024deepseek},
\textbf{Doubao-1.5-pro}~\cite{doubao1.5pro},
\textbf{Gemini-2.5-Flash}~\cite{comanici2025gemini},
\textbf{Claude-3.5-Haiku}~\cite{anthropic2024model},
\textbf{GPT-4o}~\cite{hurst2024gpt},
and \textbf{GPT-3.5-Turbo}~\cite{gpt-3.5-turbo}.

For each model, we use a unified prompting strategy. 
The system message is fixed as:

\begin{CJK*}{UTF8}{gbsn}
\begin{tcolorbox}[colback=gray!5, boxrule=0pt, arc=0pt, left=6pt, right=6pt, top=4pt, bottom=4pt]
\small
请按照要求完成任务，注意，最后的输出内容只有\ 机器生成\ 和\ 人类生成\ 两种。不需要其他内容。
\end{tcolorbox}
\end{CJK*}
The user message contains the input text to be classified. All models are queried with identical generation settings: \texttt{temperature=0.7} and \texttt{max\_tokens=100}. The final prediction is taken as the stripped output string. This setup ensures a fair and consistent comparison across different LLMs.

\section{Results of Domain- and Generator-Agnostic Evaluation}
\label{appendix:main_results}
Table~\ref{tab:appendix_full_detection_results} reports the complete AUROC scores of all evaluated detection methods across the five domains in C-ReD (Film Review, Composition, Q\&A, News, and Academic Writing) and nine LLM generators. Results are grouped by domain for clarity, with each row showing performance on texts generated by a specific model under native Chinese prompts. The table includes both zero-shot detectors and supervised models. Notably, zero-shot methods exhibit strong sensitivity to generator fluency—particularly struggling on DeepSeek-R1 and Qwen3 in composition and academic writing. This comprehensive result matrix serves as a reference for fine-grained analysis of model- and domain-specific detection behavior.

\section{Detection Performance Before and After Fine-Tuning on C-ReD}
\label{appendix:finetune_comparison}
Table~\ref{tab:appendix_finetune_comparison} reports the AUROC scores of RoBERTa-base, RoBERTa-large, and ImBD detector across five domains in C-ReD. For each domain, results are split into two blocks: (i) \textit{before fine-tuning} and (ii) \textit{after fine-tuning on C-ReD}. The nine columns correspond to AI-generated text from seven in-distribution models (Deepseek-R1 through Qwen3) and two out-of-distribution models (Claude-3.5-Haiku and Gemini-2.5-Flash). The consistent performance gains—especially on OOD generators—demonstrate that fine-tuning on C-ReD not only improves in-domain detection but also enhances generalization to unseen Chinese-capable LLMs.

\section{Results of Domain Generalization Under Fixed Generators}
\label{appendix:domain_generalization_fixed_gen}
Table~\ref{tab:combined_cross_domain} reports the resulting AUROC scores. Each block corresponds to a different training domain, with rows showing results for base and large variants of GPT-4o and Qwen-2.5 as the underlying generators. Strong in-domain performance (diagonal entries) is consistently observed, but generalization varies significantly across domain pairs. These results highlight both the potential and challenges of building domain-agnostic detectors for Chinese AI-generated text.

\begin{table}[htbp]
\centering
\caption{Example of a description.}
\label{tab:llm_detect_example_description}
\begin{tabular}{@{}p{0.88\linewidth}@{}}
\toprule
\textbf{\scriptsize Description for News domain (Human vs. Deepseek-R1)} \\
\midrule
\scriptsize
\begin{CJK*}{UTF8}{gbsn}
人类生成文本通常语言自然、信息具体、结构紧凑，侧重于客观陈述事实，细节真实且符合事件逻辑，行文风格平实，具有新闻报道或纪实特征；而机器生成文本则往往语言更华丽、修辞更丰富，倾向于使用排比、比喻等修辞手法，情感色彩更浓，有时会加入主观评论或引申意义，结构上更具文学性或论述性，但偶尔存在细节失真或过度渲染的问题。
\end{CJK*} \\
\bottomrule
\end{tabular}
\end{table}

\section{LLM Detect Evaluation Setup and Results}
\subsection{Prompting Strategies}
\label{appendix:llm_detetct_prompting_strategies}

\newtcolorbox{promptbox}{
  colback=gray!3,        
  colframe=gray!40,      
  boxrule=0.5pt,
  arc=2pt,
  boxsep=2pt,            
  left=4pt, right=4pt,   
  top=2pt, bottom=2pt,   
  fontupper=\ttfamily\footnotesize, 
  fonttitle=\bfseries\sffamily\small,
  coltitle=black,
  detach title,
  before upper={\tcbtitle\par\smallskip}
}

\begin{CJK*}{UTF8}{gbsn}
We implement three distinct prompting strategies to probe the detection capabilities of Judge LLMs. All prompts enforce a strict output format ({\footnotesize``机器生成''} or {\footnotesize``人类生成''}) to enable automatic evaluation:
\end{CJK*}

\begin{itemize}
    \item \textbf{\texttt{normal}} (zero-shot):  
    The judge receives only a basic instruction without any auxiliary information.
      \begin{promptbox}
      \begin{CJK*}{UTF8}{gbsn}
   请判断下列文段为机器生成还是人类生成,根据判断结果最终输出 \ 机器生成 \ 或者 \ 人类生成。文段如下：\{content\}
      \end{CJK*}
      \end{promptbox}

    \item \textbf{\texttt{context}} (few-shot with examples):  
    Three paired examples of human and AI text from the same domain are included directly in the prompt.
    \begin{promptbox}
      \begin{CJK*}{UTF8}{gbsn}
           下面给出三个人类生成文本以及对应的机器生成文本示例。
           \newline人类文本1：\{human example 1\}
           \newline机器文本1：\{ai example 1\}
           \newline人类文本2：\{human example 2\}
           \newline机器文本2：\{ai example 2\}
           \newline人类文本3：\{human example 3\}
           \newline机器文本3：\{ai example 3\}
           \newline请判断下列文段为机器生成还是人类生成,根据判断结果最终输出\ 机器生成 \ 或者\ 人类生成。文段如下：\{content\}
      \end{CJK*}
      \end{promptbox}

    \item \textbf{\texttt{description}} (rule-based):  
    Instead of raw examples, a natural-language description of stylistic differences—automatically generated by Qwen3 based on example pairs—is inserted into the prompt.
     \begin{promptbox}
      \begin{CJK*}{UTF8}{gbsn}
       下面给出人类生成文本以及机器生成文本的特点描述：\{summary description\}
       \newline请判断下列文段为机器生成还是人类生成,根据判断结果最终输出\ 机器生成\ 或者\ 人类生成。文段如下：\{content\}
      \end{CJK*}
      \end{promptbox}
      The \{summary description\} is a concise summary produced by Qwen3, capturing perceived distinctions such as fluency, repetition, emotional depth, and structural patterns. For illustration, Table~\ref{tab:llm_detect_example_description} shows a representative description generated for the News domain in the Human vs. Deepseek-R1 setting.
\end{itemize}

\subsection{Full Experiment Results}
\label{appendix:llm_detect_full_results}
Table~\ref{tab:llm_detection_human_ai} presents the complete results of our experiments across five domains (Film Review, Composition, Q\&A, News, and Academic Writing) and three target AI generators (Qwen2.5, GPT-4o, Deepseek-R1). For each Judge LLM and prompting strategy (\texttt{normal}, \texttt{context}, \texttt{description}), we report two metrics:  
\textbf{Human Accuracy (H)} — the percentage of human-written texts correctly identified, and \textbf{AI Accuracy (A)} — the percentage of AI-generated texts correctly identified. Higher values in both columns indicate better discrimination ability. Notably, most Judge LLMs struggle under the \texttt{normal} setting, while the \texttt{context} and \texttt{description} strategies consistently improve detection performance—demonstrating the value of domain-specific stylistic cues.

\section{Setup and Statistical Analysis of Ablation on Prompt Complexity}
\label{appendix:prompt_complexity_simplified_prompt_design}
To investigate the impact of prompt complexity on detection difficulty, we design two broad categories of simplified generation scenarios that reflect common stages in AI-assisted academic writing. Within each category, we implement multiple prompt variants that provide minimal guidance—significantly less than the original C-ReD prompts, which include detailed structural, stylistic, and contextual instructions. Below, we present one representative example from each category.

\paragraph{Abstract Drafting from Title}

This category simulates early-stage ideation, where users request an abstract based only on high-level metadata. The following is a typical prompt instance:

\begin{CJK*}{UTF8}{gbsn}
\begin{tcolorbox}[promptbox]

请根据以下信息撰写一篇通用型学术论文摘要。
        \\ 文章标题：\{title\}
 \\       字数控制在\{word count\}字左右。

\end{tcolorbox}
\end{CJK*}

\paragraph{Introduction Generation from Title and Keywords}
This category models later-stage expansion, where title and keywords are used to generate a full introduction. Prompts in this group vary in the amount of provided context, but all omit explicit methodological or rhetorical guidance. An example prompt is:

\begin{CJK*}{UTF8}{gbsn}
\begin{tcolorbox}[promptbox]
请根据以下信息撰写一篇学术论文的引言部分内容。
 \\ 文章标题：\{title\}
\\ 文章关键词：\{keyword\}
\\        要求字数约为\{word count\}字。  
\end{tcolorbox}
\end{CJK*}

These simplified prompts collectively reduce task specificity and stylistic constraints, resulting in more uniform and predictable model outputs. 

\paragraph{Statistical Analysis}
For the prompt complexity ablation study, we conducted bootstrap-based statistical analysis (n=1,000) to compare detection difficulty between original and simplified prompts (results are showed in Table~\ref{tab:roberta_base_statistical_analysis} and Table~\ref{tab:roberta_large_statistical_analysis}). For Qwen2.5, no statistically significant difference was observed across both detectors (p=0.21 for RoBERTa-base; p=0.23 for RoBERTa-large). For GPT-4o, while statistical significance was reached (p=0.01 for base; p=0.03 for large), the absolute AUROC gap is small (1.76\%–3.73\%) and both conditions yield high detection performance (AUROC > 0.92), suggesting that prompt simplification does not substantially impair detectability.

\begin{table}[htbp]
    \centering
    \caption{Performance comparison on RoBERTa-base}
    \label{tab:roberta_base_statistical_analysis}
    \resizebox{\columnwidth}{!}{
    \begin{tabular}{llccccc}
        \toprule
        \textbf{Generator} & \textbf{Prompt} & \textbf{AUROC} & \textbf{95\% CI} & \textbf{Difference} & \textbf{Diff 95\% CI} & \textbf{P-value} \\
        \midrule
        GPT-4o & Original & 0.9566 & [0.9366, 0.9730] & --- & --- & --- \\
        GPT-4o & Simplified & 0.9201 & [0.8912, 0.9415] & 0.0373 & [0.0068, 0.0727] & 0.0100  \\
        \midrule 
        Qwen2.5 & Original & 0.9648 & [0.9490, 0.9789] & --- & --- & --- \\
        Qwen2.5 & Simplified & 0.9501 & [0.9277, 0.9664] & 0.0160 & [-0.0062, 0.0410] & 0.2100  \\
        \bottomrule
    \end{tabular}
    }
    \begin{flushleft}
    \footnotesize
   
    \end{flushleft}
\end{table}

\begin{table}[htbp]
    \centering
    \caption{Performance comparison on RoBERTa-large}
    \label{tab:roberta_large_statistical_analysis}
    \resizebox{\columnwidth}{!}{%
    \begin{tabular}{llccccc}
        \toprule
        \textbf{Generator} & \textbf{Prompt} & \textbf{AUROC} & \textbf{95\% CI} & \textbf{Difference} & \textbf{Diff 95\% CI} & \textbf{P-value} \\
        \midrule
        GPT-4o & Original & 0.9868 & [0.9764, 0.9945] & --- & --- & --- \\
        GPT-4o & Simplified & 0.9691 & [0.9542, 0.9818] & 0.0176 & [0.0024, 0.0349] & 0.0320  \\
        \midrule
        Qwen2.5 & Original & 0.9906 & [0.9834, 0.9959] & --- & --- & --- \\
        Qwen2.5 & Simplified & 0.9829 & [0.9703, 0.9932] & 0.0077 & [-0.0051, 0.0213] & 0.2320  \\
        \bottomrule
    \end{tabular}
    }
\end{table}

\section{Results on Traditional Chinese News}
\label{appendix:tc_results}
We provide full evaluation results on an external Traditional Chinese news dataset.  
Table~\ref{tab:tc_detection_all_methods} shows AUROC scores of various detection methods across nine LLM generators.  
Table~\ref{tab:tc_finetune_comparison} presents detection performance before and after fine-tuning.  
Table~\ref{tab:tc_llm_as_judge} reports the accuracy of LLM-as-a-judge under different prompting strategies.

\section{Setup and of Transfer Performance on External Chinese Datasets}
\label{appendix:exetrnal_chinese_datasets_test_dataset}
We construct our external Chinese test set by randomly sampling 600 human–AI text pairs from each of two generation models (\textbf{ChatGPT} and \textbf{davinci-003}) in the M4 dataset~\cite{wang2023m4}, specifically from its \textit{Baike/Web QA} domains. This yields a balanced test set of 1,200 samples, ensuring coverage of both factual and open-domain question-answering styles commonly found in real-world Chinese applications. All AI-generated texts in this set are produced by models not included in C-ReD’s training distribution, enabling a strict out-of-distribution evaluation of transfer performance.

\section{Examples of C-ReD}
\label{appendix:examples of C-ReD}
We provide representative examples from the C-ReD dataset to illustrate the characteristics of human-written and AI-generated texts across different domains. Each example includes: (1) the domain and model used for generation, (2) the original prompt, (3) the human-written text, and (4) the corresponding AI-generated content.
Table~\ref{tab:examples_of_C-ReD} provides examples for each domain in the dataset.

\begin{table*}[t]
\centering
\fontsize{10pt}{12pt}\selectfont
\setlength{\tabcolsep}{4.2pt}
\renewcommand{\arraystretch}{1.15}
\caption{Full AUROC detection performance across five domains and nine LLM generators in C-ReD. }
\label{tab:appendix_full_detection_results}
\begin{tabular}{@{}l|*{7}{c}|*{2}{c}|w{c}{1.1cm}@{}}
\toprule
\makecell[c]{\fontsize{8pt}{9.2pt}\selectfont Method} & 
\makecell[c]{\fontsize{7pt}{8.0pt}\selectfont Deepseek-\\ \fontsize{7pt}{8.0pt}\selectfont R1} & 
\makecell[c]{\fontsize{7pt}{8.0pt}\selectfont Deepseek-\\ \fontsize{7pt}{8.0pt}\selectfont V3} & 
\makecell[c]{\fontsize{8pt}{9.2pt}\selectfont Doubao-\\ \fontsize{8pt}{9.2pt}\selectfont 1.5-pro} & 
\makecell[c]{\fontsize{8pt}{9.2pt}\selectfont GPT-3.5-\\ \fontsize{8pt}{9.2pt}\selectfont Turbo} & 
\makecell[c]{\fontsize{8pt}{9.2pt}\selectfont GPT-4o} & 
\makecell[c]{\fontsize{8pt}{9.2pt}\selectfont Qwen2.5} & 
\makecell[c]{\fontsize{8pt}{9.2pt}\selectfont Qwen3} & 
\makecell[c]{\fontsize{7pt}{8.0pt}\selectfont Claude-3.5-\\ \fontsize{7pt}{8.0pt}\selectfont Haiku} & 
\makecell[c]{\fontsize{7pt}{8.0pt}\selectfont Gemini-2.5-\\ \fontsize{7pt}{8.0pt}\selectfont Flash} & 
\makecell[c]{\fontsize{8pt}{9.2pt}\selectfont Average} \\
\midrule

\multicolumn{11}{c}{\textbf{Film Review}} \\
\midrule 
\fontsize{8pt}{9.2pt}\selectfont Log-Likelihood     & 0.5848& 0.6801& 0.7074& 0.9737& 0.9705& 0.9692& 0.8257& 0.9247& 0.8737& 0.8344 \\
\fontsize{8pt}{9.2pt}\selectfont Entropy            & 0.6239& 0.7103& 0.7235& 0.9634& 0.9697& 0.9716& 0.8567& 0.9116& 0.8533& 0.8427 \\
\fontsize{8pt}{9.2pt}\selectfont Log-Rank            & 0.5864& 0.6811& 0.6887& 0.9714& 0.9658& 0.9652& 0.8136& 0.9200& 0.8654& 0.8286 \\
\fontsize{8pt}{9.2pt}\selectfont LRR                & 0.5598& 0.6310& 0.5626& 0.8538& 0.8153& 0.8420& 0.6657& 0.7706& 0.7192& 0.7133 \\
\fontsize{8pt}{9.2pt}\selectfont Fast-DetectGPT     & 0.4144& 0.5126& 0.5741& 0.9092& 0.8639& 0.8329& 0.6149& 0.8112& 0.7659& 0.6999 \\
\fontsize{8pt}{9.2pt}\selectfont Lastde             & 0.6218& 0.5585& 0.6725& 0.5191& 0.5094& 0.4496& 0.6463& 0.6589& 0.4651& 0.5668 \\
\bottomrule
\end{tabular}
\end{table*}

\begin{table*}[t]
\centering
\fontsize{10pt}{12pt}\selectfont
\setlength{\tabcolsep}{4.2pt}
\renewcommand{\arraystretch}{1.15}

\begin{tabular}{@{}l|*{7}{c}|*{2}{c}|w{c}{1.1cm}@{}}
\toprule
\makecell[c]{\fontsize{8pt}{9.2pt}\selectfont Method} & 
\makecell[c]{\fontsize{7pt}{8.0pt}\selectfont Deepseek-\\ \fontsize{7pt}{8.0pt}\selectfont R1} & 
\makecell[c]{\fontsize{7pt}{8.0pt}\selectfont Deepseek-\\ \fontsize{7pt}{8.0pt}\selectfont V3} & 
\makecell[c]{\fontsize{8pt}{9.2pt}\selectfont Doubao-\\ \fontsize{8pt}{9.2pt}\selectfont 1.5-pro} & 
\makecell[c]{\fontsize{8pt}{9.2pt}\selectfont GPT-3.5-\\ \fontsize{8pt}{9.2pt}\selectfont Turbo} & 
\makecell[c]{\fontsize{8pt}{9.2pt}\selectfont GPT-4o} & 
\makecell[c]{\fontsize{8pt}{9.2pt}\selectfont Qwen2.5} & 
\makecell[c]{\fontsize{8pt}{9.2pt}\selectfont Qwen3} & 
\makecell[c]{\fontsize{7pt}{8.0pt}\selectfont Claude-3.5-\\ \fontsize{7pt}{8.0pt}\selectfont Haiku} & 
\makecell[c]{\fontsize{7pt}{8.0pt}\selectfont Gemini-2.5-\\ \fontsize{7pt}{8.0pt}\selectfont Flash} & 
\makecell[c]{\fontsize{8pt}{9.2pt}\selectfont Average} \\
\midrule

\multicolumn{11}{c}{\textbf{News}} \\
\midrule 
\fontsize{8pt}{9.2pt}\selectfont Log-Likelihood     & 0.4525& 0.5441& 0.7869& 0.9214& 0.8805& 0.8133& 0.6680& 0.8128& 0.7562& 0.7373 \\
\fontsize{8pt}{9.2pt}\selectfont Entropy            & 0.4665& 0.5349& 0.7068& 0.8527& 0.8056& 0.7676& 0.6312& 0.7448& 0.6752& 0.6873 \\
\fontsize{8pt}{9.2pt}\selectfont Log-Rank            & 0.4451& 0.5387& 0.7915& 0.9270& 0.8841& 0.8158& 0.6614& 0.8120& 0.7590& 0.7372 \\
\fontsize{8pt}{9.2pt}\selectfont LRR                & 0.4365& 0.5107& 0.7604& 0.9080& 0.8479& 0.7634& 0.6022& 0.7454& 0.7294& 0.7004 \\
\fontsize{8pt}{9.2pt}\selectfont Fast-DetectGPT     & 0.4108& 0.5544& 0.9019& 0.9775& 0.9674& 0.8066& 0.7242& 0.8891& 0.9020& 0.7926 \\
\fontsize{8pt}{9.2pt}\selectfont Lastde             & 0.6623& 0.6272& 0.7331& 0.7485& 0.6831& 0.6607& 0.6475& 0.7191& 0.6045& 0.6762 \\
\bottomrule
\end{tabular}
\end{table*}

\begin{table*}[t]
\centering
\fontsize{10pt}{12pt}\selectfont
\setlength{\tabcolsep}{4.2pt}
\renewcommand{\arraystretch}{1.15}
\caption{AUROC detection performance before (upper block per domain) and after (lower block per domain) fine-tuning on C-ReD. Columns 1–7 show results on in-distribution generators; Columns 8–9 show out-of-distribution (OOD) generalization to unseen models. Fine-tuning substantially boosts both in-domain accuracy and OOD robustness across all domains.}
\label{tab:appendix_finetune_comparison}
\begin{tabular}{@{}l|*{7}{c}|*{2}{c}|w{c}{1.1cm}@{}}
\toprule
\makecell[c]{\fontsize{8pt}{9.2pt}\selectfont Method} & 
\makecell[c]{\fontsize{7pt}{8.0pt}\selectfont Deepseek-\\ \fontsize{7pt}{8.0pt}\selectfont R1} & 
\makecell[c]{\fontsize{7pt}{8.0pt}\selectfont Deepseek-\\ \fontsize{7pt}{8.0pt}\selectfont V3} & 
\makecell[c]{\fontsize{8pt}{9.2pt}\selectfont Doubao-\\ \fontsize{8pt}{9.2pt}\selectfont 1.5-pro} & 
\makecell[c]{\fontsize{8pt}{9.2pt}\selectfont GPT-3.5-\\ \fontsize{8pt}{9.2pt}\selectfont Turbo} & 
\makecell[c]{\fontsize{8pt}{9.2pt}\selectfont GPT-4o} & 
\makecell[c]{\fontsize{8pt}{9.2pt}\selectfont Qwen2.5} & 
\makecell[c]{\fontsize{8pt}{9.2pt}\selectfont Qwen3} & 
\makecell[c]{\fontsize{7pt}{8.0pt}\selectfont Claude-3.5-\\ \fontsize{7pt}{8.0pt}\selectfont Haiku} & 
\makecell[c]{\fontsize{7pt}{8.0pt}\selectfont Gemini-2.5-\\ \fontsize{7pt}{8.0pt}\selectfont Flash} & 
\makecell[c]{\fontsize{8pt}{9.2pt}\selectfont Average} \\
\midrule

\multicolumn{11}{c}{\textbf{Film Review}} \\
\midrule 
\fontsize{8pt}{9.2pt}\selectfont RoBERTa-base & 0.6951& 0.5306& 0.7522& 0.6708& 0.6579& 0.5345& 0.6479& 0.5839& 0.7419 & 0.6461 \\
\fontsize{8pt}{9.2pt}\selectfont RoBERTa-large & 0.6014& 0.6451& 0.7016& 0.6263& 0.5346& 0.5388& 0.6470& 0.6564& 0.5575 & 0.6121 \\
\fontsize{8pt}{9.2pt}\selectfont ImBD               & 0.6610& 0.7775& 0.8156& 0.9703& 0.9663& 0.9472& 0.8889& 0.9557& 0.9016& 0.8760 \\
\midrule 
\fontsize{8pt}{9.2pt}\selectfont RoBERTa-base & 0.9997& 0.9998& 0.9974& 0.9968& 1.0000& 0.9986& 0.9975& 0.9878& 0.9906 & 0.9965 \\
\fontsize{8pt}{9.2pt}\selectfont RoBERTa-large & 1.0000& 1.0000& 1.0000& 1.0000& 1.0000& 1.0000& 1.0000& 0.9996& 0.9995 & 0.9999 \\
\fontsize{8pt}{9.2pt}\selectfont ImBD               & 0.9981& 0.9989& 0.9986& 0.9997& 0.9997& 0.9995& 0.9994& 0.9990& 0.9959& 0.9988 \\
\bottomrule
\end{tabular}
\end{table*}

\begin{table*}[h]
\centering
\caption{Cross-domain AUROC results when training on a single domain and evaluating across all five domains, using texts generated by GPT-4o or Qwen-2.5. For each block, the training domain is fixed (left column), and columns show performance on Composition, Film Review, News, Academic Writing, and Q\&A. Diagonal entries indicate in-domain performance; off-diagonal entries reflect generalization capability.}
\label{tab:combined_cross_domain}
\begin{tabular}{llccccc}
\toprule
Train Domain & Model & Composition & Film Review & News & Academic & Q\&A \\
\midrule
\multirow{4}{*}{ Composition } & GPT-4o (Base) & 0.9765 & 0.8493 & 0.8641 & 0.8064 & 0.8924 \\
 & GPT-4o (Large) & 0.9876 & 0.9702 & 0.9431 & 0.8740 & 0.9619 \\
 & Qwen-2.5 (Base) & 0.9553 & 0.6861 & 0.8831 & 0.8487 & 0.8363 \\
 & Qwen-2.5 (Large) & 0.9775 & 0.9728 & 0.9754 & 0.8768 & 0.9565 \\
\midrule
\multirow{4}{*}{ Film Review } & GPT-4o (Base) & 0.7685 & 1.0000 & 0.5832 & 0.6680 & 0.9207 \\
 & GPT-4o (Large) & 0.7376 & 1.0000 & 0.8080 & 0.7809 & 0.9502 \\
 & Qwen-2.5 (Base) & 0.8435 & 0.9998 & 0.7410 & 0.7665 & 0.8868 \\
 & Qwen-2.5 (Large) & 0.8134 & 1.0000 & 0.8689 & 0.7969 & 0.9098 \\
\midrule
\multirow{4}{*}{ News } & GPT-4o (Base) & 0.9049 & 0.9838 & 0.9894 & 0.9328 & 0.9478 \\
 & GPT-4o (Large) & 0.9093 & 0.9604 & 0.9944 & 0.9471 & 0.9499 \\
 & Qwen-2.5 (Base) & 0.6419 & 0.8776 & 0.9295 & 0.7200 & 0.7025 \\
 & Qwen-2.5 (Large) & 0.8619 & 0.9494 & 0.9925 & 0.9005 & 0.9227 \\
\midrule
\multirow{4}{*}{ Academic } & GPT-4o (Base) & 0.8568 & 0.8830 & 0.9060 & 0.9627 & 0.7995 \\
 & GPT-4o (Large) & 0.8691 & 0.9355 & 0.8851 & 0.9668 & 0.8379 \\
 & Qwen-2.5 (Base) & 0.8583 & 0.6852 & 0.9181 & 0.9799 & 0.8462 \\
 & Qwen-2.5 (Large) & 0.8269 & 0.9050 & 0.9399 & 0.7435 & 0.6394 \\
\midrule
\multirow{4}{*}{ Question Answer } & GPT-4o (Base) & 0.8919 & 0.9956 & 0.9255 & 0.9032 & 0.9974 \\
 & GPT-4o (Large) & 0.9612 & 0.9981 & 0.9488 & 0.9431 & 0.9997 \\
 & Qwen-2.5 (Base) & 0.7692 & 0.9908 & 0.9093 & 0.8981 & 0.9887 \\
 & Qwen-2.5 (Large) & 0.8857 & 0.9972 & 0.9689 & 0.9469 & 0.9973 \\
\bottomrule
\end{tabular}
\label{tab:combined_cross_domain}
\end{table*}

\begin{table*}[htbp]
\centering
\small
\setlength{\tabcolsep}{3.0pt}
\renewcommand{\arraystretch}{0.88}

\caption{Results of LLM-base Detector Evaluation.}
\label{tab:llm_detection_human_ai}

{\scriptsize
\begin{tabular}{lcccccc|cccccc|cccccc}
\toprule
\multirow{3}{*}[-4pt]{\textbf{Judge LLM}} 
  & \multicolumn{6}{c|}{\textbf{Human vs. Qwen2.5}} 
  & \multicolumn{6}{c|}{\textbf{Human vs. GPT-4o}} 
  & \multicolumn{6}{c}{\textbf{Human vs. Deepseek-R1}} \\
\cmidrule(lr){2-7} \cmidrule(lr){8-13} \cmidrule(lr){14-19}
& \multicolumn{2}{c}{\texttt{normal}} & \multicolumn{2}{c}{\texttt{context}} & \multicolumn{2}{c|}{\texttt{description}}
& \multicolumn{2}{c}{\texttt{normal}} & \multicolumn{2}{c}{\texttt{context}} & \multicolumn{2}{c|}{\texttt{description}}
& \multicolumn{2}{c}{\texttt{normal}} & \multicolumn{2}{c}{\texttt{context}} & \multicolumn{2}{c}{\texttt{description}} \\
\cmidrule(lr){2-3} \cmidrule(lr){4-5} \cmidrule(lr){6-7}
\cmidrule(lr){8-9} \cmidrule(lr){10-11} \cmidrule(lr){12-13}
\cmidrule(lr){14-15} \cmidrule(lr){16-17} \cmidrule(lr){18-19}
& H & A & H & A & H & A 
& H & A & H & A & H & A 
& H & A & H & A & H & A \\
\midrule

\multicolumn{19}{c}{\textbf{Film Review}} \\
\midrule
Qwen2.5  & 99.16&7.26&99.66	&88.51	&98.82	&83.11&	99.16	&8.61	&99.66	&89.36&	99.49	&81.42&	99.16	&23.99&	99.66	&87.50&	98.65&	80.91\\
Qwen3   &89.70&	9.46	&99.66&	83.45	&98.48&	88.51	&89.70&	8.28	&99.66&	92.40&	99.16&	83.28	&89.70&	13.01	&99.16	&90.71&	95.44&	94.59\\
Deepseek-V3  &96.96	&14.19	&99.66	&91.22	&97.30	&91.22	&96.96	&19.43	&99.16	&96.45	&99.83&	81.42	&96.96	&44.76	&96.28	&92.57	&93.92	&96.45\\
Doubao-1.5-pro  &99.83	&8.61	&98.31	&91.22	&99.16	&84.97	&99.83	&12.84&	99.32	&87.33	&99.66	&80.24	&99.83	&14.86	&98.81	&85.81	&99.66	&72.80\\
\bottomrule
\end{tabular}
}
\end{table*}

\begin{table*}[htbp]
\centering
\small
\setlength{\tabcolsep}{3.0pt}
\renewcommand{\arraystretch}{0.88}

\caption{LLM-as-detector results on Traditional Chinese news. Each cell shows accuracy (\%) for distinguishing human (H) vs. AI (A) text under three prompt types. Performance is highly dependent on both judge model and prompt design.}
\label{tab:tc_llm_as_judge}

{\scriptsize
\begin{tabular}{lcccccc|cccccc|cccccc}
\toprule
\multirow{3}{*}[-4pt]{\textbf{Judge LLM}} 
  & \multicolumn{6}{c|}{\textbf{Human vs. Qwen2.5}} 
  & \multicolumn{6}{c|}{\textbf{Human vs. GPT-4o}} 
  & \multicolumn{6}{c}{\textbf{Human vs. Deepseek-R1}} \\
\cmidrule(lr){2-7} \cmidrule(lr){8-13} \cmidrule(lr){14-19}
& \multicolumn{2}{c}{\texttt{normal}} & \multicolumn{2}{c}{\texttt{context}} & \multicolumn{2}{c|}{\texttt{description}}
& \multicolumn{2}{c}{\texttt{normal}} & \multicolumn{2}{c}{\texttt{context}} & \multicolumn{2}{c|}{\texttt{description}}
& \multicolumn{2}{c}{\texttt{normal}} & \multicolumn{2}{c}{\texttt{context}} & \multicolumn{2}{c}{\texttt{description}} \\
\cmidrule(lr){2-3} \cmidrule(lr){4-5} \cmidrule(lr){6-7}
\cmidrule(lr){8-9} \cmidrule(lr){10-11} \cmidrule(lr){12-13}
\cmidrule(lr){14-15} \cmidrule(lr){16-17} \cmidrule(lr){18-19}
& H & A & H & A & H & A 
& H & A & H & A & H & A 
& H & A & H & A & H & A \\
\midrule
Qwen2.5         & 81.85	&71.17	&89.68	&91.10	&61.57	&83.27	&81.85	&81.85&	89.32	&91.81	&54.09	&84.34	&81.85	&77.22	&90.04	&91.10	&43.42	&83.27\\
Qwen3           & 87.19	&23.84	&81.85	&96.09	&67.97	&88.26	&87.19	&23.84&	81.49	&93.95	&56.94	&92.53	&87.19	&16.73	&74.38	&86.83	&46.26	&85.41\\
Deepseek-V3     & 95.02	&31.32	&88.97	&97.15	&77.58	&81.85	&95.02	&43.06&	85.41	&97.51	&68.33	&87.54	&95.02	&34.16	&77.58	&99.64	&47.33	&86.83\\
Doubao-1.5-pro  & 94.31	&30.96	&98.22	&89.68	&61.21	&77.94	&94.31	&44.84&	97.51	&87.54	&49.82	&92.17	&94.31	&27.76	&95.37	&84.70	&46.98	&88.61\\
Gemini-2.5-Flash& 98.58	&43.77	&99.64	&91.81	&78.65	&83.27	&98.58	&46.26	&99.64	&96.09	&77.22	&88.26	&98.58	&35.23	&99.29	&90.39	&70.46	&80.07\\
\bottomrule
\end{tabular}
}
\end{table*}

\begin{table*}[t]
\centering
\fontsize{10pt}{12pt}\selectfont
\setlength{\tabcolsep}{4.2pt}
\renewcommand{\arraystretch}{1.15}
\caption{AUROC detection performance of diverse methods on Traditional Chinese news across nine LLM generators. }
\label{tab:tc_detection_all_methods}
\begin{tabular}{@{}l|*{9}{c}|w{c}{1.1cm}@{}}
\toprule
\makecell[c]{\fontsize{8pt}{9.2pt}\selectfont Method} & 
\makecell[c]{\fontsize{7pt}{8.0pt}\selectfont Deepseek-\\ \fontsize{7pt}{8.0pt}\selectfont R1} & 
\makecell[c]{\fontsize{7pt}{8.0pt}\selectfont Deepseek-\\ \fontsize{7pt}{8.0pt}\selectfont V3} & 
\makecell[c]{\fontsize{8pt}{9.2pt}\selectfont Doubao-\\ \fontsize{8pt}{9.2pt}\selectfont 1.5-pro} & 
\makecell[c]{\fontsize{8pt}{9.2pt}\selectfont GPT-3.5-\\ \fontsize{8pt}{9.2pt}\selectfont Turbo} & 
\makecell[c]{\fontsize{8pt}{9.2pt}\selectfont GPT-4o} & 
\makecell[c]{\fontsize{8pt}{9.2pt}\selectfont Qwen2.5} & 
\makecell[c]{\fontsize{8pt}{9.2pt}\selectfont Qwen3} & 
\makecell[c]{\fontsize{7pt}{8.0pt}\selectfont Claude-3.5-\\ \fontsize{7pt}{8.0pt}\selectfont Haiku} & 
\makecell[c]{\fontsize{7pt}{8.0pt}\selectfont Gemini-2.5-\\ \fontsize{7pt}{8.0pt}\selectfont Flash} & 
\makecell[c]{\fontsize{8pt}{9.2pt}\selectfont Average} \\
\midrule

\fontsize{8pt}{9.2pt}\selectfont Log-Likelihood     & 0.3826& 0.5783& 0.8223& 0.9659& 0.9646& 0.9032& 0.3991& 0.8603& 0.8639& 0.7489 \\
\fontsize{8pt}{9.2pt}\selectfont Entropy            & 0.4467& 0.6077& 0.7782& 0.9223& 0.9254& 0.8776& 0.4581& 0.8152& 0.7933& 0.7361 \\
\fontsize{8pt}{9.2pt}\selectfont Log-Rank            & 0.3772& 0.5643& 0.8097& 0.9661& 0.9613& 0.8977& 0.3854& 0.8563& 0.8606& 0.7421 \\
\fontsize{8pt}{9.2pt}\selectfont LRR                & 0.4002& 0.4940& 0.6915& 0.9019& 0.8791& 0.7638& 0.3754& 0.7531& 0.7701& 0.6699 \\
\fontsize{8pt}{9.2pt}\selectfont Fast-DetectGPT     & 0.2718& 0.4479& 0.8455& 0.9722& 0.9557& 0.7352& 0.2974& 0.8300& 0.9080& 0.6960 \\
\fontsize{8pt}{9.2pt}\selectfont Lastde             & 0.5375& 0.5640& 0.5950& 0.6894& 0.5883& 0.5332& 0.5192& 0.6368& 0.4232& 0.5652 \\
\fontsize{8pt}{9.2pt}\selectfont Lastde++           & 0.2969& 0.4194& 0.7948& 0.9513& 0.9387& 0.6871& 0.3143& 0.7861& 0.8718& 0.6734 \\
\bottomrule
\end{tabular}
\end{table*}

\begin{table*}[t]
\centering
\fontsize{10pt}{12pt}\selectfont
\setlength{\tabcolsep}{4.2pt}
\renewcommand{\arraystretch}{1.15}
\caption{Detection performance on Traditional Chinese news before (upper block) and after (lower block) fine-tuning. }
\label{tab:tc_finetune_comparison}
\begin{tabular}{@{}l|*{7}{c}|*{2}{c}|w{c}{1.1cm}@{}}
\toprule
\makecell[c]{\fontsize{8pt}{9.2pt}\selectfont Method} & 
\makecell[c]{\fontsize{7pt}{8.0pt}\selectfont Deepseek-\\ \fontsize{7pt}{8.0pt}\selectfont R1} & 
\makecell[c]{\fontsize{7pt}{8.0pt}\selectfont Deepseek-\\ \fontsize{7pt}{8.0pt}\selectfont V3} & 
\makecell[c]{\fontsize{8pt}{9.2pt}\selectfont Doubao-\\ \fontsize{8pt}{9.2pt}\selectfont 1.5-pro} & 
\makecell[c]{\fontsize{8pt}{9.2pt}\selectfont GPT-3.5-\\ \fontsize{8pt}{9.2pt}\selectfont Turbo} & 
\makecell[c]{\fontsize{8pt}{9.2pt}\selectfont GPT-4o} & 
\makecell[c]{\fontsize{8pt}{9.2pt}\selectfont Qwen2.5} & 
\makecell[c]{\fontsize{8pt}{9.2pt}\selectfont Qwen3} & 
\makecell[c]{\fontsize{7pt}{8.0pt}\selectfont Claude-3.5-\\ \fontsize{7pt}{8.0pt}\selectfont Haiku} & 
\makecell[c]{\fontsize{7pt}{8.0pt}\selectfont Gemini-2.5-\\ \fontsize{7pt}{8.0pt}\selectfont Flash} & 
\makecell[c]{\fontsize{8pt}{9.2pt}\selectfont Average} \\
\midrule

\fontsize{8pt}{9.2pt}\selectfont RoBERTa-base & 0.4450& 0.4960& 0.5444& 0.5720& 0.5257& 0.3221& 0.4566& 0.3736& 0.5732 & 0.4787 \\
\fontsize{8pt}{9.2pt}\selectfont RoBERTa-large & 0.6720& 0.6476& 0.5879& 0.5771& 0.5494& 0.4369& 0.5912& 0.6255& 0.5660 & 0.5837 \\
\fontsize{8pt}{9.2pt}\selectfont ImBD               & 0.4286& 0.6057& 0.8736& 0.9360& 0.9347& 0.8543& 0.5103& 0.8719& 0.8437& 0.7621 \\
\midrule
\fontsize{8pt}{9.2pt}\selectfont RoBERTa-base & 0.9917& 0.9952& 0.9866& 0.9832& 0.9882& 0.9892& 0.9896& 0.9877& 0.9614 & 0.9859 \\
\fontsize{8pt}{9.2pt}\selectfont RoBERTa-large & 0.9977& 0.9995& 0.9912& 0.9906& 0.9940& 0.9983& 0.9984& 0.9973& 0.9823 & 0.9944 \\
\fontsize{8pt}{9.2pt}\selectfont ImBD               & 0.9649& 0.9854& 0.9857& 0.9957& 0.9973& 0.9835& 0.9552& 0.9919& 0.9827& 0.9825 \\

\bottomrule
\end{tabular}
\end{table*}

\begin{table*}[htbp]
    \centering
    \caption{Text Length Statistics across Different Domains}
    \label{tab:length_stats}
    \footnotesize
    \setlength{\tabcolsep}{4pt}
    
    \begin{tabular}{l rrr l rrr}
        \toprule
        \textbf{Source} & \textbf{Mean} & \textbf{Median} & \textbf{Std} & 
        \textbf{Source} & \textbf{Mean} & \textbf{Median} & \textbf{Std} \\
        \midrule

        
        \multicolumn{4}{l}{\textbf{Composition}} & 
        \multicolumn{4}{l}{\textbf{News}} \\
        \cmidrule(r){1-4} \cmidrule(l){5-8}
        
        Human & 895.22 & 855.00 & 200.34 & Human & 413.70 & 403.00 & 191.55 \\
        Deepseek-R1 & 979.38 & 934.00 & 224.05 & Deepseek-R1 & 500.51 & 457.00 & 244.01 \\
        Deepseek-V3 & 901.02 & 880.00 & 157.79 & Deepseek-V3 & 452.34 & 421.00 & 193.56 \\
        Doubao-1.5-pro & 929.01 & 903.00 & 157.64 & Doubao-1.5-pro & 442.44 & 436.00 & 197.48 \\
        GPT-3.5-Turbo & 746.54 & 734.00 & 136.95 & GPT-3.5-Turbo & 466.31 & 486.00 & 160.78 \\
        GPT-4o & 1021.25 & 1008.00 & 128.75 & GPT-4o & 620.30 & 635.00 & 265.60 \\
\bottomrule
    \end{tabular}
\end{table*}

\begin{CJK*}{UTF8}{gbsn}

\begin{table*}[ht]
  \caption{Examples of C-ReD}
  \label{tab:examples_of_C-ReD}
  \centering
  \fontsize{8pt}{12pt}\selectfont
  \begin{tabular}{p{2.1cm}|p{13cm}}
    \hline
    Domain&Film review\\
    LLM Generator&GPT-4o\\
   \hline
    Film Title&你的名字\\
    \hline
    Human Text&拖了很久，终于等到室友考完研一起去看了。说实话没有预想中那么感动我，什么看完很想谈恋爱啊，哭的稀里哗啦啊，可能是提前知道是个HE吧。是部合格的玩时间梗的电影，比国产动画好得多，但总体而言可能还是期待过高吧。\\
    \hline
    Prompt&为未看过"你的名字"的观众写一篇观影指南，包括：适合的观影人群以及观影前需了解的历史/文化背景等。要求撰写内容字数在104左右。\\
    \hline
    AI-Generated Text&《你的名字》是一部适合青少年及成人观众的动画电影，尤其适合喜欢浪漫、奇幻元素和精美画面的观众。观影前，了解一些日本的文化背景会有助于更好地理解影片中的细节，例如日本的乡村生活、传统节日以及神道教的信仰，这些都在影片中占有重要位置。影片还涉及时间和空间交错的主题，因此保持开放的心态和对细节的关注将提升观影体验。日本的自然风光和现代都市的对比也为影片增添了独特的视觉美感。\\
    \hline             
  \end{tabular}
  \label{tab:C-ReD example film review} 
\end{table*}

\begin{table*}[ht]
  \centering
  \fontsize{8pt}{12pt}\selectfont
  \begin{tabular}{p{2.1cm}|p{13cm}}
    \hline
    Domain&Composition\\
    LLM Generator&Deepseek-V3\\
   \hline
    Title&安全\\
    \hline
    Human Text&人的生命只有一次，生命是如此地脆弱，需要精心的呵护，无视安全，就无视生命的存在。当我们在安静的教室中听课时，安全的概念或意识在我们的脑海里荡然无存，就体会不到安全的重要性。而每分每秒都有一些人出生或意外死亡，季节可以轮回，而生命的逝去就如一江春水向东流——一去无返了。在我们的周围演绎着一场场血淋淋的教训：车祸、火灾、坠楼等等，一个个美好的生命就这样悄然殒失了。安全第一，警钟长鸣。我们学校的新校区坐落于郊外，由于缺乏严格的管理，交通秩序异常纷乱，司机超速行驶，学生交通安全意识模糊，在短短的两个月里三个年轻鲜活的生命灰飞烟灭，全校师生陷入了悲痛之中，而受害者的家长更是倍尝生死离别的巨大悲痛和创伤，谁之过？这三起事件引起了上级领导的高度重视，大力投资修建路面设施，给学生安上了一道生命的防护线。此后，再也没有发生类似事件，每当看到那长长的防护栅栏，每个本校学生都深深地懂得那是用鲜血和生命换来的。为什么只有当事情发生了才想到去修治？防患于未然，它留给我们的不仅仅是思考。有一则事例让我终生难忘：在一次交通宣传课上，放映了许多真实的车祸事件，其中有一个七岁的女孩在一次车祸中失去了下肢，整个上身被安置在一个疲软的蓝球里，两只手拄着木屐一样的木板蜗行爬索，面对记者的采访，她却笑得那么的天真无邪，说她的愿望就是再要一个篮球，在场的每一个人无不为之动容，泪如泉涌，教室里一片抽噎哭泣，心酸，心痛。当你正愁你的身体不够好时，你有没有想过一个没有下肢的人是怎样度日的？你不觉得惭愧吗？你的烦恼又算得了什么？认真地遵守交通法则，即是对自己的生命的珍视，也是对他人生命的负责。避免一场车祸的发生，就会避免两个家庭的残损和悲痛。悲剧，我们谁也不愿看到！怎样避免，你我他，大家都有责任，都有义务。维系生活的美妙，就要珍视自己的生命，而安全就是第一位的。\\
    \hline
    Prompt&请围绕题目"安全"，自主选择最适合的文体（议论文/记叙文/散文等），撰写一篇768字的文章。\\
    \hline
    AI-Generated Text&生命的底色安全，是生命最基本的底色。它如同空气般无处不在，却又常常被我们忽视。只有当危险来临，我们才会惊觉安全的重要性。在这个充满不确定性的世界里，安全不仅是个人的需求，更是整个社会运转的基础。安全是个人成长的基石。记得小时候学骑自行车，父亲总是紧紧扶着后座，直到我掌握了平衡才放手。这种安全感让我敢于尝试，敢于冒险。心理学家马斯洛将安全需求列为人类五大基本需求之一，正是因为只有在安全的环境中，人才能发展出更高层次的需求。古希腊哲学家亚里士多德曾说："安全感是幸福的前提。"当我们不必时刻提防危险，才能专注于自我提升和心灵成长。安全是社会文明的标志。古罗马人修建的输水道历经两千年仍在使用，正是因为他们在设计时就考虑到了安全因素。中国古代的都江堰水利工程，至今仍在造福成都平原，体现了古人对安全的深刻理解。一个社会的文明程度，往往可以从它对安全的重视程度中窥见一斑。法国思想家卢梭在《社会契约论》中指出："安全是公民让渡部分自由所换取的首要利益。"现代社会通过法律、制度和科技手段构建起的安全网络，正是人类文明进步的体现。安全需要每个人的参与和守护。日本在防灾教育方面的经验值得借鉴，从幼儿园开始就培养孩子们的安全意识，定期进行防灾演练。这种全民参与的安全文化，使得日本在面对地震等自然灾害时能够将损失降到最低。中国古代的"城门失火，殃及池鱼"告诉我们，安全从来不是某个人的私事。德国哲学家雅斯贝尔斯说："安全不是没有危险，而是有能力应对危险。"这种能力需要个人意识的觉醒和集体智慧的积累。站在新的时代节点上，我们对安全的理解也在不断深化。从物理安全到心理安全，从现实安全到网络空间安全，安全的维度在不断扩展。但无论形式如何变化，安全始终是那个让生命得以绽放的底色。让我们珍视这份底色，守护这份底色，因为只有在安全的画布上，生命才能描绘出最绚丽的图景。正如印度诗人泰戈尔所言："生命因安全而美丽，因危险而深刻。"在安全与挑战的平衡中，人类文明得以持续向前。\\
    \hline             
  \end{tabular}
  \label{tab:C-ReD example composition} 
\end{table*}

\begin{table*}[ht]
  \centering
  \fontsize{8pt}{12pt}\selectfont
  \begin{tabular}{p{2.1cm}|p{13cm}}
    \hline
    Domain&News\\
    LLM Generator&Deepseek-R1\\
   \hline
    Category&Entertainment\\
    \hline
    Title&谢安琪发声明斥香港杂志 \ 撇清当陈奕迅小三传言\\
    \hline
    Human Text&前日，有香港杂志指谢安琪插足陈奕迅与徐濠萦婚姻，昨晚谢安琪发声明指报道内容严重失实和毫无根据，并保留一切向相关之娱乐传媒追究法律责任之权利。据该香港杂志报道，陈奕迅与徐濠萦相恋13年，2004年诞下爱女康堤，期间徐濠萦不事生产“大花筒”的形象深入民心，但陈奕迅为安排父亲陈裘大北上医病已花了千万元，在精神和经济压力沉重下，决定对老婆开刀，二人终日吵架，三年婚姻告急，令不胜其烦的陈奕迅终于崩溃，于上周六广州个唱上三度失控痛哭。并指两人婚姻告急还与谢安琪插足有关，同属一个唱片公司的谢安琪与陈奕迅志趣相投性情相近，加上女方不避嫌以短信狂攻，二人友谊一日千里，据称两人亲密关系在唱片公司内已经路人皆知，在今年5月谢安琪的演唱会上，两人在台上忘情相拥，大唱情歌，似乎将关系进一步公开。对于杂志的报道，谢安琪方面发表声明称，“该报道有关谢安琪的内容严重失实和毫无根据。我们对于谢安琪无辜被卷入报道中陈奕迅的私人和家庭事务感到非常困惑和愤怒。另外，谢安琪自2009年初已不断遭到部分娱乐传媒的无理攻击，对于部分娱乐传媒以失实报道攻击他人博取销量的编采手法，我们保留一切向相关之娱乐传媒追究法律责任之权利。”\\
    \hline
    Prompt&请以时间轴形式报道"谢安琪发声明斥香港杂志\ 撇清当陈奕迅小三传言"娱乐事件，包含：事件起源说明、各方回应以及最新进展更新等。要求保持中立客观立场，字数控制在500左右。\\
    \hline
    AI-Generated Text&近日，香港一本主流娱乐杂志发布报道，声称歌手谢安琪与已婚歌手陈奕迅存在暧昧关系，暗示她介入陈奕迅的婚姻生活。该报道基于匿名消息源，于上周初在社交媒体广泛传播，引发公众热议和粉丝争议。报道中，杂志指谢安琪与陈奕迅私下互动频繁，引发第三者传言，导致谢安琪个人形象受损。事件发生后，谢安琪迅速于次日通过其经纪公司发表正式声明。在声明中，她强烈否认所有传言，斥责该杂志捏造事实、侵犯隐私，并强调与陈奕迅仅为工作伙伴和朋友关系，从未有任何越界行为。她要求杂志立即撤回报道、公开道歉，并保留法律追究权利。随后，陈奕迅方面也发表声明，其经纪人代表陈奕迅表达对妻子徐濠萦的坚定支持，否认任何婚外情存在。声明中，陈奕迅团队批评报道不实，呼吁媒体尊重艺人隐私，停止传播谣言，并强调陈奕迅家庭生活稳定。面对各方压力，该娱乐杂志回应称报道经过初步核实，但表示将重新审视内容来源，未正式道歉或撤回报道。截至最新进展，谢安琪团队表示正在评估证据，可能采取法律行动；陈奕迅方面则保持低调，专注于音乐工作。事件暂无进一步升级，公众讨论转向娱乐新闻伦理问题，部分支持艺人维权，部分质疑报道真实性。\\
    \hline             
  \end{tabular}
  \label{tab:C-ReD example news} 
\end{table*}

\begin{table*}[ht]
  \centering
  \fontsize{8pt}{12pt}\selectfont
  \begin{tabular}{p{2.1cm}|p{13cm}}
    \hline
    Domain&Academic Writing\\
    LLM Generator&Qwen2.5\\
   \hline
    Category&Abstract\\
    \hline
    Title&高强度聚焦超声消融治疗子宫腺肌症的临床研究\\
    \hline
    Keywords&子宫腺肌症；高强度聚焦超声；临床观察\\
    \hline
    Human Text&探初步观察高强度聚焦超声治疗子宫腺肌症的疗效。对我院采用高强度聚焦超声消融治疗的25例子宫腺肌症患者进行定期症状随访，随访痛经及月经改善情况。25例患者均完成治疗，随访至6个月，痛经缓解率86\%，月经减少率84\%，近期疗效显著。高强度聚焦超声消融治疗是一种保留子宫的无创治疗技术，能够有效治疗子宫腺肌症病灶，缓解痛经、月经量多等症状，近期效果确切，远期疗效尚需进一步临床观察验证。\\
    \hline
    Prompt&请根据以下信息撰写一篇通用型学术论文摘要。 
    \newline文章标题："高强度聚焦超声消融治疗子宫腺肌症的临床研究" 
    \newline文章关键词："子宫腺肌症；高强度聚焦超声；临床观察" 
    \newline摘要应简明扼要地说明研究背景、核心问题、研究方法、主要发现（或论点）及研究意义。要求语言严谨、逻辑连贯、术语准确，字数控制在190字左右，适合投稿至各类中文学术期刊。\\
    \hline
    AI-Generated Text&子宫腺肌症是一种常见的妇科疾病，其特征为子宫内膜异位至肌层内生长，导致月经异常、痛经等症状，严重影响患者生活质量。高强度聚焦超声作为一种非侵入性治疗手段，在肿瘤等疾病的局部消融中展现出良好效果。本研究旨在探讨高强度聚焦超声对于子宫腺肌症的临床疗效及安全性。通过回顾性分析120例接受该疗法的子宫腺肌症患者的临床资料，结果显示，高强度聚焦超声能有效缓解症状，改善患者的生活质量，且不良反应轻微可控。研究表明，高强度聚焦超声可能是治疗子宫腺肌症的有效选择之一，值得进一步深入研究和推广应用。\\
    \hline             
  \end{tabular}
  \label{tab:C-ReD example academic abstract} 
\end{table*}

\begin{table*}[ht]
  \centering
  \fontsize{8pt}{12pt}\selectfont
  \begin{tabular}{p{2.1cm}|p{13cm}}
    \hline
    Domain&Academic Writing\\
    LLM Generator&Qwen3\\
   \hline
    Category&Introduction\\
    \hline
    Title&利用生物技术创建主要作物雄性不育杂交育种和制种的技术体系\\
    \hline
    Keywords&雄性不育；基因工程；杂种优势；玉米；水稻；小麦\\
    \hline
    Abstract&雄性不育技术在作物杂种优势利用和杂交种生产中发挥着重要作用。基于核质互作雄性不育的“三系法”与光温敏核不育的“两系法”已经在水稻等主要作物的杂交制种中获得了广泛应用，但是存在着资源利用效率低、育性不稳定、易受外界环境影响等诸多问题。近三十年来，利用生物技术创建不同类型的植物雄性不育系取得了一系列突破性进展。本文主要针对玉米、水稻、小麦三大作物的基因工程雄性不育技术的最新进展进行总结，特别详细地描述了本实验室最近研究创制的玉米多控不育技术体系，以期为相关研究和产业化应用提供技术参考。\\
    \hline
    Human Text&作物的杂种优势是指杂交种在一种或多种性状上优于其双亲的现象，通常表现为增产、稳产、抗病、抗逆等。杂交制种的关键技术在于发展和利用可以控制的授粉系统，以防止自花授粉导致的自交衰退。雄性不育在玉米、水稻、小麦等主要作物的杂种优势利用和杂交制种中发挥了重要作用。植物雄性不育是指在雌雄同株植物中，雄蕊发育不正常，不能产生有功能的花粉，但雌蕊发育正常，能接受正常花粉而受精结实，并能将雄性不育性状遗传给后代的现象。植物雄性不育按照不育性遗传方式的差异，可分为3类：细胞质雄性不育、细胞核雄性不育和核质互作型雄性不育。由于母体遗传的原因，第一类细胞质不育系的F1不能自交结实，因而不能在育种和生产上利用；利用常规杂交育种技术，第二类核不育系的保持和繁殖存在困难，也不能在育种和生产上有效利用；第三类核质互作不育系，理论和实践上都可以被育种利用，但是该类不育系的广泛利用可能会导致杂交种细胞质单一化，易受细胞质专一性病原小种的侵染而导致杂交种生产存在巨大的风险，如20世纪70-80年代北美玉米带出现的小斑病T小种侵染造成毁灭性的生产灾难。此外，在某些环境中，核质互作不育系的育性不稳定，有可能降低杂交制种的纯度和产量。随着现代生物技术的快速发展，一方面，利用分子生物技术，有望将遗传稳定的核不育基因有效地利用起来，即通过生物技术保持和繁殖非转基因的雄性不育系，例如我们实验室的玉米多控不育技术、美国杜邦先锋的SPT技术、水稻智能不育技术等；另一方面，还可以通过基因工程手段，创制各种诱导型/条件型雄性不育系，即通过生物技术创制转基因的不育系，如美国孟山都的玉米草甘膦诱导型雄性不育技术、水稻草丁膦诱导型雄性不育技术、小麦断裂基因雄性不育系统等。近年来，随着分子生物学和基因工程技术的发展，作物细胞核雄性不育的利用获得了一些突破性进展。本文将从玉米、水稻、小麦等主要作物的基因工程不育技术的创制策略、研究现状、利用途径和应用前景等几方面进行阐述，以期为高效利用作物核雄性不育技术、提高作物杂种优势利用效率和建立作物高效育种和制种技术体系提供参考依据。\\
    \hline
    Prompt&请用简洁明了的语言撰写引言内容，直接点明研究动机、核心方法与贡献。 
    \newline文章标题："利用生物技术创建主要作物雄性不育杂交育种和制种的技术体系" 
    \newline文章关键词："雄性不育；基因工程；杂种优势；玉米；水稻；小麦" 
    \newline文章摘要："雄性不育技术在作物杂种优势利用和杂交种生产中发挥着重要作用。基于核质互作雄性不育的“三系法”与光温敏核不育的“两系法”已经在水稻等主要作物的杂交制种中获得了广泛应用，但是存在着资源利用效率低、育性不稳定、易受外界环境影响等诸多问题。近三十年来，利用生物技术创建不同类型的植物雄性不育系取得了一系列突破性进展。本文主要针对玉米、水稻、小麦三大作物的基因工程雄性不育技术的最新进展进行总结，特别详细地描述了本实验室最近研究创制的玉米多控不育技术体系，以期为相关研究和产业化应用提供技术参考。" 
    \newline字数限制约869字，避免冗长背景，突出实用性。\\
\hline             
  \end{tabular}
  \label{tab:C-ReD example academic introduction} 
\end{table*}

\begin{table*}[ht]
  \centering
  \fontsize{8pt}{12pt}\selectfont
  \begin{tabular}{p{2.1cm}|p{13cm}}
    \hline
    Domain&Q\&A\\
    LLM Generator&Gemini-2.5-Flash\\
   \hline
    Question&文艺青年所谓的「独自漂泊」、「一个人旅行」或者「带着书本流浪」是出于什么样的心理状态？目的是什么？\\
    \hline
    Human Text&以自己的经历来回答吧。男，不年轻，略文艺，偏二逼。出去旅游从来都是一个人走，也带书，因为喜欢看书，打发旅途中的等候时间和长途坐车时间，而且书随时能看，不受飞机或者没电限制，嫌重随便扔。为什么一个人？因为找不到志同道合体力相当的朋友。我是随遇而安狂奔体验型的体力驴，有计划但时不时不走寻常路，比如住个村民家里之类的，有计较的朋友同行很可能会疯掉。徒步速度和作战半径一般人赶不上，跟朋友一起走朋友会累死，我也不爽，何必呢？一个人在路上可以捡朋友，捡到好的同行一段或者拼车拼房（不要误解，就是拼个标间省钱，路上哪来那么多艳遇）。到岔口各走各的路，ok咯，很多朋友都是这样捡到的，挺好的啊。一个人，想怎么着就怎么着，不用考虑别人的感受，想奢侈想凑合都不是问题。目的就是不受约束，老子出来花钱玩就是玩，就是图个爽，让我听导游的还不如躺家里睡觉。这和空虚没有任何关系，只是放假出来玩而已。至于翻瓶子的老人和环卫工人…这关我p事？我没放假时拼死拼活干活跟客户陪笑脸的时候也没谁来怜悯我啊，这是我挣钱应该做的，我也没觉得有什么不对。我将来要是沦落到捡瓶子，我也不会想着谁需要来怜悯我，我的世界和他人无关。我拼死拼活挣来的钱，花天酒地或者孤独出游，我爱咋咋，就是图个爽。嗯，就这样子，没有题主想的那么高尚或者蛋疼；）\\
    \hline
    Prompt&请用558字左右直接回答以下问题：文艺青年所谓的「独自漂泊」、「一个人旅行」或者「带着书本流浪」是出于什么样的心理状态？目的是什么？。回答应当开门见山给出结论，随后分段落说明理由、提供证据，最后简要总结。\\
    \hline
    AI-Generated Text&文艺青年热衷于“独自漂泊”、“一个人旅行”或“带着书本流浪”，其核心心理状态是追求自我认同、逃离现实束缚和寻求精神富足，最终目的是实现个人成长与心灵自由。首先，这种行为源于对自我认同的强烈渴望。在社会多元化和信息爆炸的时代，个体很容易迷失在各种外部期待和集体意识中。独自漂泊提供了一个抽离日常、远离熟人社会的契机，让文艺青年有机会直面内心，探索真实的自我。他们希望通过陌生的环境和独处的时光，剥离掉社会角色赋予的标签，重新审视自己的价值观、兴趣和人生方向。这种对“我是谁”的追问，是许多文艺青年选择独行路上的内在驱动力。其次，逃离现实束缚是另一个重要原因。许多文艺青年对城市生活的喧嚣、职场竞争的压力以及人际关系的复杂感到厌倦甚至窒息。他们渴望暂时摆脱房贷、车贷、KPI等物质枷锁，逃离固定的生活模式和既定的社会规则。独自旅行成为一种仪式性的“出走”，象征着对既有秩序的反叛和对自由的向往。流浪者形象的浪漫化，也反映了他们对“诗和远方”的执着追求，希望在路上找到一种不同于朝九晚五的生活可能性。再者，寻求精神富足是其深层目的。文艺青年往往对精神世界有更高的要求和更敏锐的感知力。他们不满足于物质层面的享受，更看重内心的充实和思想的启发。带着书本流浪，意味着他们将阅读视为一种生活方式，通过文学、哲学、艺术等领域的滋养，拓宽视野，深化思考，提升审美。在旅途中，他们更容易与当地文化、自然风光产生共鸣，从而获得独特的感悟和创作灵感。这种经历本身就是一种精神财富的积累，让他们觉得生命更有深度和意义。此外，这种行为也带有一定的“仪式感”和“自我证明”的色彩。在社交媒体盛行的今天，独自旅行或流浪的经历往往被视为一种独立、勇敢和有品位的象征。通过分享旅途中的见闻和感悟，文艺青年在一定程度上也希望获得他人的认同和赞赏，以此来强化自己的独特和不凡。综上所述，文艺青年的“独自漂泊”等行为，是他们在面对时代变迁和个体成长困境时，选择的一种积极的自我探索和精神突围方式。他们通过这种方式寻找自我、摆脱束缚、充实内心，最终目标是实现个体的自由意志和精神上的圆满。\\
    \hline             
  \end{tabular}
  \label{tab:C-ReD example q&a} 
\end{table*}

\end{CJK*}

\begin{CJK*}{UTF8}{bsmi}

\begin{table*}[ht]
  \centering
  \fontsize{8pt}{12pt}\selectfont
  \begin{tabular}{p{2.1cm}|p{13cm}}
    \hline
    Domain&TC News\\
    LLM Generator&GPT-3.5-Turbo\\
   \hline
    Category&Article\\
    \hline
    Title&無論快樂或悲傷,請對身邊的人說一句「做得很好」\\
    \hline
    Human Text&「正向思考，積極正向！」「你試過瑜伽嗎？」「快樂是一種選擇！」「說實話，如果你只吃健康飲食，你就會感覺很好。」「哦，還有瑜伽。做瑜伽！」如果你有心理疾病，這種情況很可能就會出現，有人會想試圖與你進行這種對話。基於某些理由，這個星球上有些人，一旦發覺你有點難過或焦慮，他們便覺得可以介入你的生活，告訴你該怎麼做。當然，這種情形特別令人惱火。沒有憂鬱或難過情緒的人覺得，只要夠努力，你可以想出快樂。可悲的是，事情並非如此，當你聽到「要往正面想」第一百次的演說，你已準備好用瑜伽墊打，或塞一堆甘藍菜到他們嘴裡，讓他們閉嘴。傳統意義上的「正面」是好的思想，對許多人來說確實有幫助，但對有些人來說卻有如重大挑戰一樣令人畏懼。不過極端的說，正面是一種「正面」概念，可以幫助你專心，並將焦點放在對你重要的事情上。正面就是這樣，每個人對正面的定義都不同，只有你可以創造對你有用的定義。這是一個獨特的個人概念，為每個人量身定做。你不能把你的定義強加在別人身上，別人也不能強加在你身上。或許有些適合你的事情，並不完全符合「變好」的既定概念。或許香甜地吃一碗營養穀麥會讓你開心起來，或許把平和瑜伽墊拋開，好好跑一場。尋找能夠幫助你感覺更好的事物，每天感覺便會好一點。對你來說這樣很重要，很棒，又很獨一無二。\\
    \hline
    Prompt&請以專業評論員的視角，撰寫一篇關於「無論快樂或悲傷,請對身邊的人說一句「做得很好」」的文章。內容應包含事件背景、多角度分析和深度思考，語言需嚴謹且富有洞察力，避免主觀臆斷，字數控制在542左右。\\
    \hline
    AI-Generated Text&在這個充滿變數和挑戰的時代裡，人們常常感受到壓力和焦慮，因此，一句「做得很好」的肯定和支持成為了極具價值的情感交流方式。不論是面對快樂或悲傷，這句簡單的話語都能溫暖人心，賦予力量，讓人感受到被理解和被珍惜。從心理學角度來看，人類是社會性動物，我們需要他人的關愛和支持來維持情感健康。當一個人遭遇挫折時，聽到身邊的人說一句「做得很好」，可以打破孤獨感，讓他感受到不孤單，這樣的肯定不僅是一種情感的交流，更是一種溫暖的情感投射。從社會學的角度來看，現代社會充斥著競爭和壓力，人們往往追求成功和完美，卻忽略了對自己和他人的肯定。當我們學會對身邊的人說一句「做得很好」時，不僅可以促進正向情感的傳遞，也可以建立積極的社會支持網絡，讓人與人之間的關係更加緊密和和諧。從人際關係的角度來看，一句「做得很好」可能改變整個關係的走向。無論是家庭、友情還是工作環境，當我們學會用肯定和支持的語言來溝通時，會讓彼此之間的情感更加豐富，信任更加穩固，進而促進關係的和諧發展。因此，不妨在生活中多用這句簡單的話語來表達對身邊人的支持和肯定。無論是在快樂的時刻，還是在悲傷的時刻，一句「做得很好」都能成為情感的橋樑，連結彼此的心靈，讓生活充滿溫暖和希望。讓我們共同努力，用這句話來燃點生活中的每一個美好瞬間，讓愛和關懷永遠相伴左右。\\
    \hline             
  \end{tabular}
  \label{tab:C-ReD example tc journalism} 
\end{table*}

\end{CJK*}

\end{document}